\documentclass[11pt]{article}

\usepackage[preprint]{acl}

\usepackage{times}
\usepackage{latexsym}

\usepackage[T1]{fontenc}

\usepackage[utf8]{inputenc}

\usepackage{microtype}

\usepackage{inconsolata}

\usepackage{graphicx}

\usepackage{algorithm}
\usepackage{algorithmic}

\usepackage{amssymb}
\usepackage{multirow}
\usepackage{url}            %
\usepackage{booktabs}       %
\usepackage{amsfonts}       %
\usepackage{nicefrac}       %
\usepackage{microtype}      %
\usepackage{xcolor}         %

\usepackage{tikzducks}
\usepackage{arydshln}
\usepackage{amsmath}
\usepackage{subcaption}
\usepackage{CJKutf8}

\newcommand{\ours}{K-Merge}
\newcommand{\oursp}{K-Merge$++$}

\newcommand{\keypoint}[1]{\noindent \textbf{#1}:}

\title{
K-Merge: Online Continual Merging of Adapters for\\ On-device Large Language Models
}
\author{
    \textbf{Donald Shenaj}$^{1,2,3}\thanks{Research completed during internship at Samsung R\&D Institute UK, while affiliated with the University of Padova.}$\quad
    \textbf{Ondrej Bohdal}$^{1}$\quad
    \textbf{Taha Ceritli}$^{1}$\quad
    \textbf{Mete Ozay}$^{1}$\\
    \textbf{Pietro Zanuttigh}$^{3}$\quad
    \textbf{Umberto Michieli}$^{1}$\\[0.5em]
    \textsuperscript{\rm 1}Samsung R\&D Institute UK\quad
    \textsuperscript{\rm 2}University of Pisa%
    \quad
    \textsuperscript{\rm 3}University of Padova%
}

\begin{document}

\maketitle

\begin{abstract}
On-device deployment of Large Language Models (LLMs) frequently leverages Low-Rank Adapters (LoRAs) to support diverse downstream tasks under tight resource constraints. To address the limited storage capacity of mobile devices, recent works have explored model merging techniques to fuse multiple LoRAs into a single one. In practice, however, LoRAs are often delivered incrementally, as users request support for new tasks (e.g., novel problem types or languages). 
This scenario introduces a new challenge: on-device online continual merging, where the objective is to incorporate new LoRAs while preserving the performance on previously supported tasks. 
In this paper, we propose a data-free and computationally efficient strategy for selecting and merging LoRAs when a new one becomes available, assuming the device can store only a limited number of adapters. Extensive experiments across real-world tasks demonstrate the superiority of our approach compared to alternative strategies while adhering to the storage budget and compute limitations of on-device settings. The project page is available at: \url{https://donaldssh.github.io/K-Merge}. 
\end{abstract}

\section{Introduction}
\label{sec:intro}

Large Language Models (LLMs) are powerful general-purpose models that can be adapted to a wide range of problem types in many languages, including question answering \cite{sticha2024qa}, translation \cite{zhu2023multilingual}, summarization \cite{liu2023learning}, text rewriting \cite{shu2024rewritelm}, and grammar correction \cite{severyn2021grammar}. 
Due to their substantial parameter count \cite{zhao2023survey,minaee2024large}, fine-tuning these models for task-specific applications (e.g., a problem type in a given language) is commonly approached via parameter-efficient tuning methods \cite{ding2022delta,han2024parameter}.
Among these, Low-Rank Adapters (LoRAs) \cite{hu2021lora} are especially popular: they insert lightweight trainable modules into the base model and update only these modules during fine-tuning, freezing the  LLM weights. %
This enables deploying a single base LLM and dynamically loading small LoRAs to support various tasks.
\begin{figure}[t]
    \centering
    \includegraphics[width=\linewidth]{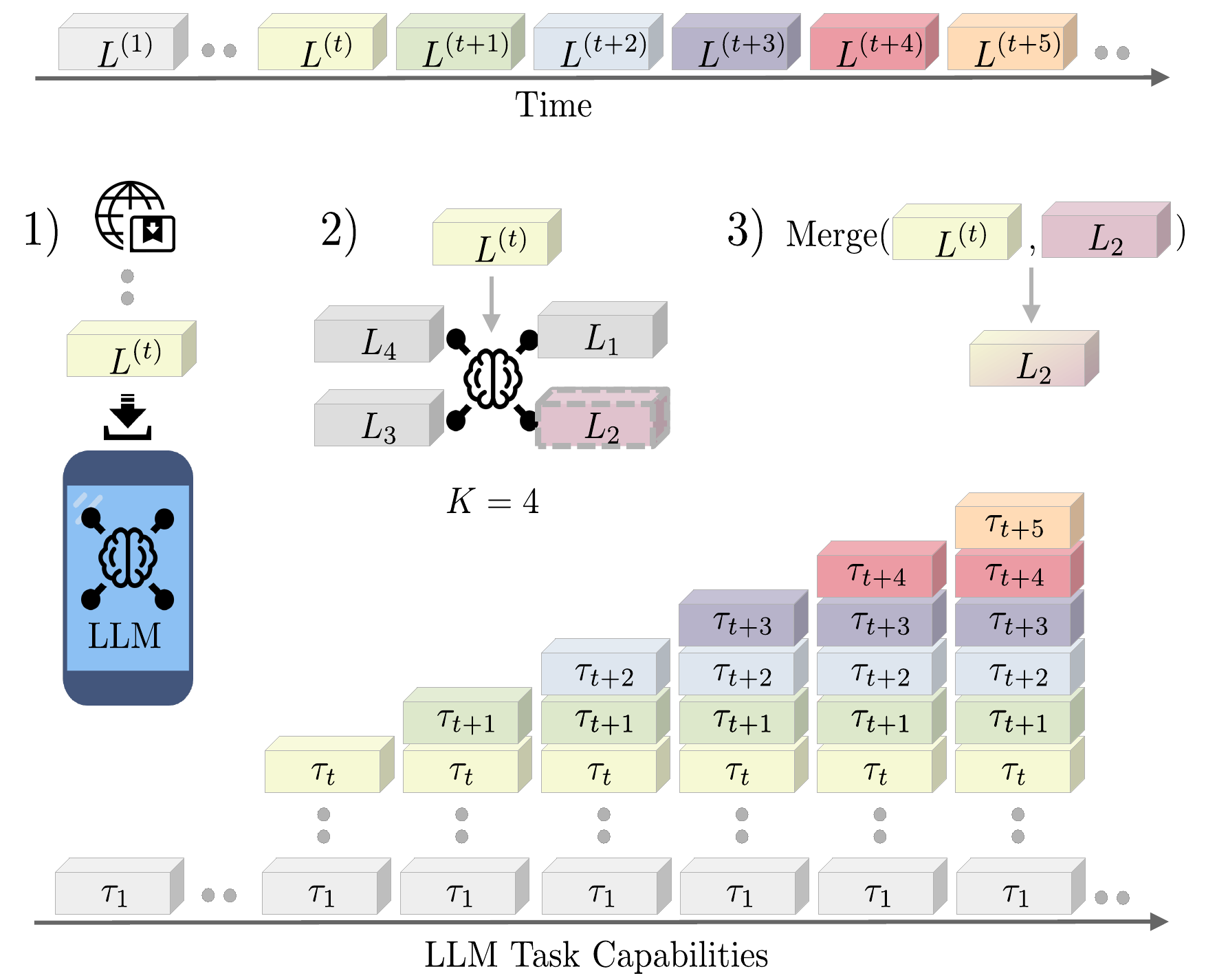}
    \caption{Online continual merging of adapters for on-device LLMs. 
     Each adapter corresponds to a specific task $\tau_t$ (e.g., 
     a specific problem type in a selected language). The objective is to increase the LLM capabilities over time without storing all adapters, but rather using a budget of $K$ adapters (e.g., $K\!=\!4$ here). The key steps are:
     1) The device downloads new adapters over time. 
     2) The system selects the most similar stored adapter to the new one. 
     3) The system updates the selected stored adapter by merging it with the new adapter.
    }
    \label{fig:teaser}
\end{figure}
A growing use case for LLMs is their deployment on mobile and edge devices \cite{dhar2021ondevice}, driven by benefits such as improved privacy, lower latency, reduced server load, and offline access. 
However, the size of LLMs suitable for on-device deployment is typically limited to 1--2B parameters, much smaller than their server-scale counterparts, e.g., 70--400B models \cite{dubey2024llama}. 
In such settings, prompting alone often results in subpar performance. 
Instead, task-specific LoRAs are commonly employed to recover acceptable accuracy \cite{burke2023aicore,gunter2024apple,ceritli2025hydraopt,bohdal2025data}, offering a practical way to extend functionality without retraining the base model.

While a single LoRA may suffice for one task, a key challenge arises in real-world usage where LoRAs arrive incrementally, as users request new functionality over time. 
However, on-device storage is limited, making it infeasible to retain a separate LoRA for every task. 
One promising solution is to construct multi-task adapters by merging existing LoRAs \cite{yang2024model}. 
Several merging strategies have been proposed, from linear combinations with fixed or learned coefficients \cite{wortsman2022model,ilharco2022editing} to more advanced approaches such as TIES \cite{yadav2024ties} and DARE \cite{yu2024language}.

A remaining question then is how to integrate each new LoRA given only the existing adapters and a limited number of storage slots. 
Crucially, the original single-task LoRAs and any training data are unavailable when a new LoRA arrives. Moreover, future task requests are \textit{a-priori} unknown. 
Therefore, the device must continually update its set of LoRAs in a lightweight, data-free, and storage-aware manner, preserving existing capabilities while integrating new ones.

In this work, we formalize this setting as online continual merging of LoRAs under a storage budget, and introduce a novel method tailored to this challenge. 
Our approach identifies which existing adapter is most similar to the incoming LoRA, %
decides whether to allocate a new slot (until the storage budget is hit) or to merge adapters using an efficient data-free strategy (see
Fig.~\ref{fig:teaser} for an overview of our pipeline).

Our contributions are summarized as follows: 
\begin{enumerate}
    \item We introduce a new and practical setting for online continual merging of adapters in resource-constrained on-device LLMs under a storage budget. 
    \item We propose a lightweight, data-free merging strategy that identifies suitable adapters for merging using information from the history of merges and employs dynamic weighting to balance old and new capabilities.
    \item We conduct extensive evaluation on real-world tasks representative of mobile device usage: %
    our approach achieves strong performance under realistic constraints.
\end{enumerate}

\section{Related Work}
\label{sec:related}

\keypoint{Parameter-Efficient Fine-Tuning (PEFT)} PEFT techniques adapt pre-trained models by training only a small subset of parameters \cite{ding2022delta,han2024parameter}. Among these, LoRAs \cite{hu2021lora} have emerged as a widely adopted approach. 
LoRA inserts two low-rank matrices into each target layer of the model and updates only these during fine-tuning, keeping the base model weights frozen. 
This yields strong performance while significantly reducing memory and compute overhead \cite{mao2025survey}.
Subsequent work has focused on improving LoRA’s efficiency and expressiveness, e.g., AdaLoRA \cite{zhang2023adalora}, Delta-LoRA \cite{zi2023delta}, DoRA \cite{liu2024dora}, VeRA \cite{kopiczko2023vera}, and Tied-LoRA \cite{renduchintala2024tied}. 
However, all these methods assume a static task setting and do not address continual merging of adapters under resource constraints.

\keypoint{Model Merging} 
Model merging aims to combine multiple models or adapters trained for different tasks into a single multi-task model. Early approaches relied on simple weight averaging \cite{wortsman2022model,ilharco2022editing}, but more sophisticated methods have since emerged \cite{huang2023lorahub,xiao2024lm,gauthier2024merging,bohdal2025compositional,hammoud2024model}. Two widely used techniques are TIES \cite{yadav2024ties}, which resets negligible parameters and resolves sign conflicts before merging, and DARE/DARE-TIES \cite{yu2024language}, which selectively drops and rescales parameter deltas.
While many model merging techniques target full-model weight merging, recent works have focused on merging LoRA adapters specifically \cite{stoica2025model,zhao2025merging,shenaj2024lora,ceritli2025hydraopt}, often enabling more modular and efficient adaptation. 
Other efforts address merging in continual learning scenarios \cite{marczak2024magmax,sokar2025continual}, particularly for discriminative tasks such as image classification \cite{tang2025merging,colemanadaptive} and retrieval \cite{dziadzio2024merge}.
TIME \cite{dziadzio2024merge} sequentially trains and merges multimodal experts using tailored initialization and merging strategies for continual integration. 
In contrast, OPCM \cite{tang2025merging} merges models without retraining by projecting updates orthogonally to prior ones, aligning better with our setting. However, its performance deteriorates as the number of merges increases.
Our work instead focuses on on-device continual merging of LoRA adapters for generative text tasks, enabling dynamic integration of new functionality over time under strict compute and memory constraints.

\keypoint{On-device LLMs}
Standard LLMs often consist of several billion parameters and require GPU-based training and inference \cite{borzunov2024distributed}. 
Since most users lack access to such infrastructure, LLM services are commonly deployed via remote servers. However, privacy concerns arise when user data must be transmitted externally--particularly in applications involving personal conversations or sensitive content.
As a result, there is growing interest in on-device LLMs to perform inference locally on mobile or edge hardware \cite{dhar2021ondevice}. 
Applications of interest include summarizing private chats \cite{gliwa2019samsum} or generating personalized responses \cite{jandaghi2023faithful}, while ensuring data remains local. 
To meet the resource demands of mobile platforms, LLMs are typically downsized to 1–2B parameters and are often further optimized via quantization or pruning \cite{zhu2024survey}.
Recent open-source models designed for on-device deployment include Llama-3.2-1B \cite{dubey2024llama} and Qwen-2.5-1.5B \cite{qwen2,qwen2.5}.
These models deliver strong language capabilities within the computational and memory limitations of modern smartphones.

\section{Problem Formulation}
\label{sec:problem_formulation}
Training multi-task or multi-lingual LoRAs from scratch is impractical due to its lack of flexibility, as it requires access to all datasets and necessitates retraining whenever a new capability is added. Hence we assume only single-task LoRAs are sent to the device.
While storage is limited, storing a small number of (possibly multi-tasking) adapters is typically feasible. Formally, we assume the availability of a storage budget for storing a collection of $\leq K$ LoRAs on the device, $\mathcal{L}=\{L_i\}_{i=1}^K$. 

In the conventional static setting, where all single-task LoRAs $L^{(t)}, \forall t$ are available simultaneously, multi-task adapters can be constructed by grouping LoRAs into $K$ clusters based on task similarity and merging each group into a shared multi-task LoRA $L_i, i=1,\ldots, K$. 

In many real-world scenarios, however, LoRAs arrive incrementally over time, giving rise to the online continual merging setup considered here.
As a result, single-task LoRAs are not accessible all at once, and merging must be performed iteratively, based solely on the incoming LoRA and the current set of stored adapters, as described next.

The merging happens on-device and not on the server, because it would be impractical and inefficient to store a copy of the deployed LoRAs on both the server and the device.

\subsection{Online Continual Merging}

At each discrete time step $t>0$, the current on-device collection $\mathcal{L}^{(t-1)}$ contains $|\mathcal{L}^{(t-1)}|\leq K$ LoRAs, where $|\cdot|$ indicates the cardinality of the set. A new single-task LoRA $L^{(t)}$ arrives, trained to support a new user-selected task $\tau_t$. 
The system must incorporate $L^{(t)}$ into $\mathcal{L}^{(t-1)}$ to obtain an updated set $\mathcal{L}^{(t)}$ whose LoRAs preserve performance across all tasks (Fig.~\ref{fig:teaser}).
We initialize $\mathcal{L}^{(0)}=\emptyset$ and denote the set of all tasks seen at the end of step $t$ by $\mathcal{T}^{(t)}=\{\tau_i\}_{i=1}^t$.

A system tackling the online continual merging setup must:
\begin{enumerate}
   
    \item Update the on-device collection whenever a new adapter arrives, by performing one of the following two actions:
    \begin{enumerate}
        \item Selecting an adapter $L_c \in \mathcal{L}^{(t-1)}$ to be merged with $L^{(t)}$
        and replace it in $\mathcal{L}^{(t-1)}$ to obtain:
        \begin{equation}
        \label{eq:merge_1}
            \mathcal{L}^{(t)} = \{\mathcal{L}^{(t-1)} \setminus L_c\} \cup \{\mathrm{merge}(L_c, L^{(t)})\},
        \end{equation}  

        where the $\mathrm{merge}(\cdot)$ operation returns the merged LoRA and should be data-free and computationally-efficient, suitable for on-device execution (see Sec.~\ref{sec:method:history});
        
        \item Allocating a new slot, i.e., adding the incoming LoRA to the current collection by:
        \begin{equation}
        \label{eq:addition}
        \mathcal{L}^{(t)} = \{ L^{(t)} \} \cup \mathcal{L}^{(t-1)}.
        \end{equation}
        Note that this operation is only available if there is still storage available, i.e., if $|\mathcal{L}^{(t-1)}|<K$.
    \end{enumerate}
    
    \item Define a map $\Theta^{(t)}$ such that, whenever an input sample $x$ of task $\tau_{i}\in\mathcal{T}^{(t)}$ arrives, it identifies the corresponding adapter's identifier, $\hat{c}=\Theta^{(t)}(i)$, to use for inference,
    \begin{equation}
    \label{eq:gamma}
    \Theta^{(t)}\!:\!i\!\in\!\{\!1,\ldots,\!|\mathcal{T}^{(t)}|\!\}\!\to\!\hat{c}\in\!\{\!1,\ldots,\!|\mathcal{L}^{(t-1)}|\!\}.
    \end{equation}
    Then, the adapter $L_{\hat{c}}$ is loaded from storage and plugged into the model to perform inference.
    
    \item Maintain strong performance on all previous tasks in $\mathcal{T}^{(t)}$.
\end{enumerate}

\subsection{Evaluation Protocol and Aggregate Score}

In the considered experimental setting, each task $\tau_i$ is characterized by a problem type and a language.
To evaluate performance, we construct a benchmark spanning $\alpha$ problem types and $\beta$ languages, totaling $\gamma= \alpha \cdot \beta$ distinct tasks for which we obtain $\gamma$ single-task LoRAs. 
Tasks arrive at the device in random order, one per time step. 
At each time step $t$, the system updates the collection of on-device LoRAs $\mathcal{L}^{(t)}$, and the performance is measured on all the tasks seen so far $\mathcal{T}^{(t)}$. 
In our setup, $t=1,\ldots,\gamma$, however, $\gamma$ is unknown to the device at any time $t$.

Each task $\tau_i$ is associated with a dataset $\mathcal{D}_{\tau_i}$ and an evaluation metric $M_{\tau_i}$ that takes as input a LoRA and provides as output the performance on the associated dataset.
Since different problem types may be evaluated by different metrics that operate on different scales,
we normalize performance on each task by comparing it to the performance of a single-task LoRA trained directly on that task. 
Let $L$ denote a generic adapter. For each task $\tau_i \in \mathcal{T}^{(t)}$, where $t>0$, let $M_{\tau_i}(L; \mathcal{D}_{\tau_i})$ be the score obtained by the LLM using $L$, 
and $M_{\tau_i}(L^{(i)}; \mathcal{D}_{\tau_i})$ the single-task performance, i.e., the score obtained by the LLM using $L^{(i)}$. 
We define the normalized aggregate score $S^{(t)}$ as follows:

\begin{equation}
    \label{eq:score}
    S^{(t)} = \frac{1}{|\mathcal{T}^{(t)}|}\sum_{\tau_i\in \mathcal{T}^{(t)}} \frac{M_{\tau_i}(L; \mathcal{D}_{\tau_i})}{M_{\tau_i}(L^{(i)}; \mathcal{D}_{\tau_i})}.
\end{equation}

This score quantifies how closely the performance of a generic LoRA $L$ aligns with that of single-task LoRAs $L^{(i)}$. 
In our experiments, we evaluate LoRAs $L\! \in \! \mathcal{L}^{(t)}$; however, the proposed score formulation is generic and allows the evaluation of any given LoRAs $L$.
For instance, $\mathcal{L}^{(t)}$ could contain either single-task LoRAs
or merged LoRAs
. Additionally, $L$ is initialized to zeroes in the zero-shot case where no LoRAs are used.

\subsection{Task Identification}
In the settings we consider, it is typically easy to identify which task should be performed. The user can explicitly select the task or language, e.g., via UI elements. Alternatively,  the task can be specified in the user prompt (e.g., ``translate this sentence'', or ``summarize this passage''). In such cases, determining the appropriate adapter is straightforward as we can identify the task via matching to selected keywords. When task identity is not directly stated, lightweight task classifiers can be used. Since routing is not related to our contribution, we model this step abstractly and assume access to a perfect classifier for simplicity. This assumption isolates the evaluation of the merging method itself without conflating it with the performance of a separate routing module.

\section{Proposed Method}
\label{sec:method}

\begin{figure*}[t]
    \centering
    \includegraphics[width=0.98\linewidth]{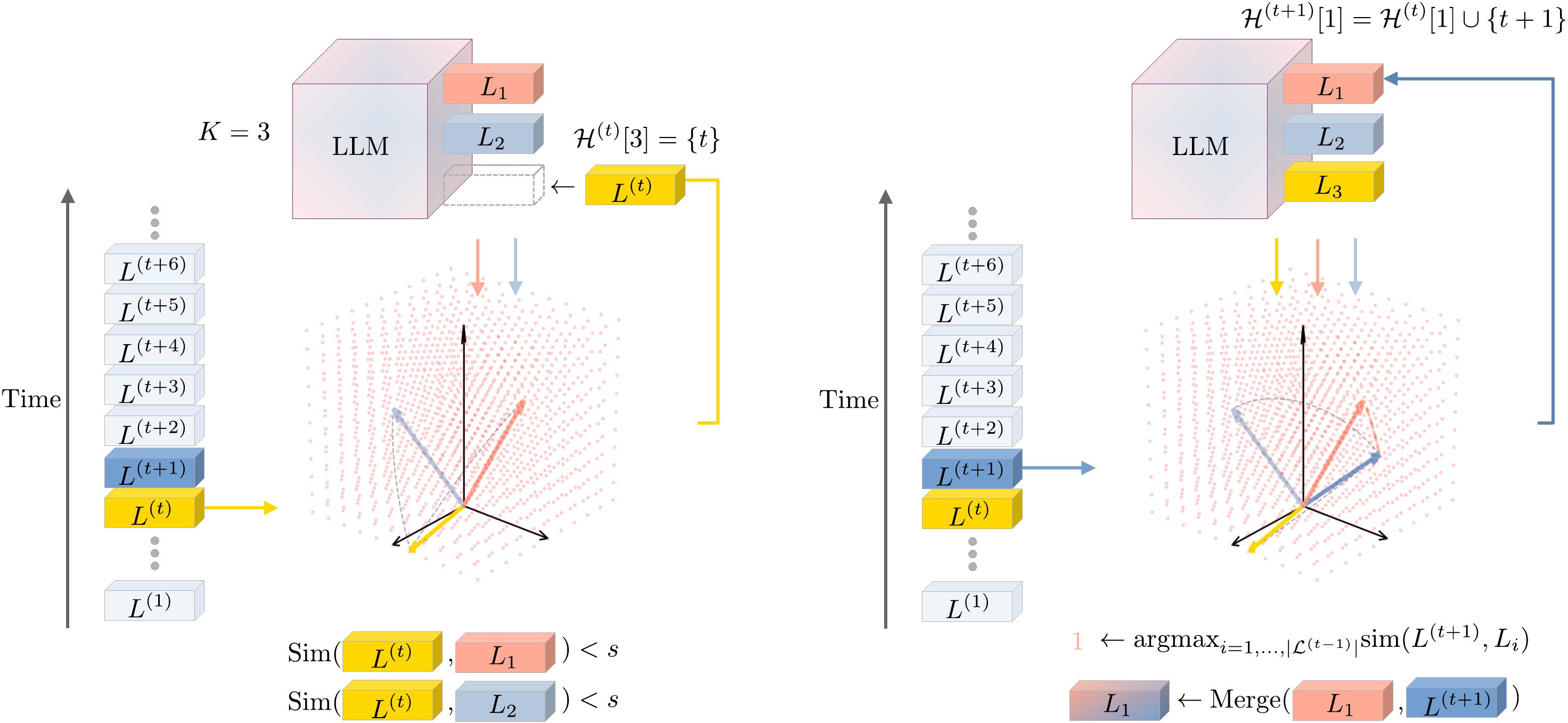}
    \caption{{K-Merge$++$ outline.} When the storage budget is not fulfilled %
    and a new LoRA $L^{(t)}$ is downloaded, it is compared with the LoRAs in $\mathcal{L}^{(t-1)}$ using Eq.~\eqref{eq:sim}.  If similarity is smaller than the threshold $s$, $L^{(t)}$ initializes a new cluster; otherwise, it gets merged with the closest LoRA following Eq.~\eqref{eq:closest}. After the budget limit has been reached, new LoRAs can only be merged.}
    \label{fig:method}
\end{figure*}

To address the challenges of online continual merging, we propose a lightweight, data-free method that supports on-device continual merging within a given storage budget. 
 An overview is shown in Fig.~\ref{fig:method}.
At the core of our method, there are two key components:
1) \textbf{Similarity-based clustering} (Sec.~\ref{sec:method:similarity}): we identify the most suitable stored adapter to merge with for an incoming LoRA, based on an efficient similarity metric without using any training data.
2) \textbf{History-aware merging} (Sec.~\ref{sec:method:history}): we merge adapters using a weighted combination scheme that leverages information from the previous adapter merges.

These components are integrated into a decision-making framework to determine whether the new LoRA should be merged into an existing adapter or allocated to a new slot (subject to the storage budget). 
We introduce two variants:
\begin{itemize}
    \item \textbf{K-Merge}, a lightweight solution that merges each new LoRA with its nearest stored adapter unless there is space available to store it separately.
    \item \textbf{K-Merge$++$},  an improved variant that selectively merges based on a similarity threshold, preserving space for future, more diverse LoRAs.
\end{itemize}

\subsection{Cluster Assignment by Similarity Scoring}
\label{sec:method:similarity}

At the core of both variants is the ability to match the incoming LoRA $L^{(t)}$ to the most compatible stored adapter among $L_i \in \mathcal{L}^{(t-1)}, \forall i$ via similarity scoring.

Each LoRA $L$ consists of a pair of low-rank update matrices $(A^{n, p}, B^{n, p})$ for each transformer layer $n \in \mathcal{N}$ and projection type ${p \in \mathcal{P} = \{\texttt{key}, \texttt{query}, \texttt{value}, \texttt{output}\}}$. 
Then, we perform the following steps:
\begin{enumerate}
    \item Flatten each update matrix $\Delta W^{n,p}=B^{n,p}A^{n,p}$ into a vector via the flattening operation $f(\cdot)$. 
    \item Compute the cosine similarity between the flattened vector corresponding to the incoming LoRA and that of all the stored LoRAs, and average over all %
    layers and projections:
\begin{equation}
\label{eq:sim}
    \text{sim}(L^{(t)}, L_i) = \frac{\sum\limits_{n\in \mathcal{N},p\in \mathcal{P}}{\text{sim}(L^{(t)n,p}, L_i^{n,p})}}{|\mathcal{N}| \cdot|\mathcal{P}|},
\end{equation}
\begin{multline*}
{\text{sim}(L^{(t)n,p}, L_i^{n,p})}\\ = \text{cos}(f(\Delta W^{(t)n,p}), f(\Delta W_i^{n,p})).
\end{multline*}

\item Determine the index $c$ of the most-similar  LoRA by:
\begin{equation}
\label{eq:closest}
c = \text{argmax}_{i=1,\ldots,
|\mathcal{L}^{(t-1)}|
} \text{sim}(L^{(t)}, L_i).
\end{equation}

\end{enumerate}

\subsection{Iterative History-aware Model Merging} %
\label{sec:method:history}

\keypoint{History Design} To retrieve the correct LoRA at the inference stage, we allocate some small storage (negligible in practice) to collect the time-varying history of merges as a map $\mathcal{H}^{(t)}$ with up to $K$ keys, each linked to a set of task indices. 
The history is initialized 
as $\mathcal{H}^{(0)}=\emptyset$.

\keypoint{Save Incoming LoRA to Storage}
Whenever an incoming adapter $L^{(t)}$ is selected to be stored on the on-device collection, a new key is added to the history and
the task index is added to the value set, as:
\begin{equation}
\label{eq:simple_addition}
    \mathcal{H}^{(t)}[|\mathcal{L}^{(t-1)}|+1] = \{t\}.
\end{equation}

\keypoint{History-aware Merging}
In case the incoming LoRA is selected to be merged with the most similar adapter $L_c$, we implement a history-aware merging. %
We implement the merge operation via a running average formulation, defined by:

\begin{equation}
\label{eq:merge}
    \mathrm{merge}(L_c,L^{(t)}) = \frac{L^{(t)} + |\mathcal{H}^{(t-1)}[c]| \cdot L_c}{|\mathcal{H}^{(t-1)}[c]| + 1} ,
\end{equation}
and the history of merges is updated accordingly as follows:
\begin{equation}
\label{eq:history}
    \mathcal{H}^{(t)}[c] = \mathcal{H}^{(t-1)}[c] \cup \{t\}.
\end{equation}

This formulation enables efficient incremental merging without requiring storage of all LoRAs in each cluster. A key benefit of the formulation is also that it is order-invariant as all the LoRAs have the same contribution to the merged multi-tasking LoRA.

\subsection{K-Merge and K-Merge$++$}

We now describe the two variants of our overall system, both designed to manage a fixed adapter budget of $K$ slots. In contrast, existing merging methods would use only one slot, while na\"ive storing of all single-task LoRAs would use one slot for each LoRA.

\keypoint{\ours} 
The basic strategy merges $L^{(t)}$ into its closest stored adapter $L_c$ using Eq.~\eqref{eq:merge}, unless fewer than $K$ adapters are currently stored, in which case $L^{(t)}$ is stored directly. 
This approach is simple and efficient, but has a key limitation: if early incoming LoRAs are similar, it may exhaust the storage budget prematurely by storing redundant adapters, thus leading to a sub-optimal clustering of adapters at later stages.

\keypoint{\oursp}
To mitigate this issue, we introduce a more advanced approach relying on a similarity threshold $s$. A new LoRA $L^{(t)}$ is merged into $L_c$ only when they are sufficiently similar to each other, i.e., $\text{sim}(L^{(t)}, L_c) \geq s$. 
Otherwise, it is stored separately if storage is still available. 
Once the storage is full, the merging is performed regardless of $s$.
The threshold $s$ is estimated empirically from an auxiliary dataset of LoRAs trained on held-out tasks. 
Specifically, we compute all pairwise similarities across LoRAs and set $s$ to the median.
This variant preserves storage for more diverse future LoRAs, improving adaptability across diverse deployment scenarios. 
Its only overhead is the need to estimate $s$.
The procedure is detailed in Algorithm~\ref{alg:continual_thr}.

\begin{algorithm}[tb]
\caption{\oursp}
\label{alg:continual_thr}
\begin{algorithmic}[1] %
\STATE \textbf{Parameters}: Storage budget $K$, similarity threshold $s$.
\STATE Initialize on-device collection by $\mathcal{L}^{(0)}=\emptyset$.
\STATE Initialize history by $\mathcal{H}^{(0)}=\emptyset$.
\WHILE{new LoRA $L^{(t)}$ is received, $t>0$}
\IF {$|\mathcal{L}^{(t-1)}| > 0$}
\STATE Find index $c$ of most similar LoRA via Eq.~\eqref{eq:closest}.
\ENDIF
\IF {$|\mathcal{L}^{(t-1)}|=K$ or ($|\mathcal{L}^{(t-1)}|>0$ and $\text{sim}(L^{(t)}, L_c) \ge s$)}
\STATE Update collection $\mathcal{L}^{(t)}$ via Eqs.~\eqref{eq:merge_1}, \eqref{eq:merge}.
\STATE Update history $\mathcal{H}^{(t)}$ via Eq.~\eqref{eq:history}.
\ELSE
\STATE Update collection via Eq.~\eqref{eq:addition}.
\STATE Update history $\mathcal{H}^{(t)}$ via Eq.~\eqref{eq:simple_addition}.
\ENDIF
\ENDWHILE
\STATE \textbf{return} $\mathcal{L}^{(t)}$
\end{algorithmic}
\end{algorithm}

\subsection{Inference Stage}
We implement the mapping described in Eq.~\eqref{eq:gamma} as an inverse function of the history $\mathcal{H}^{(t)}$.
To process a query input sample $x$ belonging to task $\tau_i \in \mathcal{T}^{(t)}$, we look for the key $\hat{c}\in\{1,\ldots,|\mathcal{L}^{(t-1)}|\}$ whose value set contains $i$, i.e., $i \in \mathcal{H}^{(t)}[\hat{c}]$.
Then, we load the adapter $L_{\hat{c}}$ from storage, plug it into the model, and perform inference.

\section{Experimental Evaluation}
\label{sec:results}
\subsection{Experimental Details}

\keypoint{Tasks and Metrics} 
We evaluate our method across a diverse set of tasks, each defined by a combination of a problem type and a target language, and each associated with a specific evaluation metric.
We consider $\alpha = 5$ problem types: \textit{Smart Reply}, \textit{Summarization}, \textit{Tone Adjustment}, \textit{Question Answering}, and \textit{Grammar Correction}.
Each of these problem types is instantiated in $\beta = 8$ languages: English, Spanish, French, German, Italian, Korean, Japanese, and Chinese. As metrics, we report F-0.5 for Grammar Correction, F-1 for Question Answering, weighted ROUGE \cite{zhang2021dataset} for Smart Reply, and ROUGE-L for both Text Summarization and Tone Adjustment.

\keypoint{Datasets} 
We use established benchmarks for each task:
1) \textit{Smart Reply:} Persona-Chat Synthetic \cite{jandaghi2023faithful}. 
2) \textit{Summarization:} SAMSum \cite{gliwa2019samsum}. 
3) \textit{Tone Adjustment:} Sound Natural \cite{einolghozati2020sound} modified via a publicly available model fine-tuned for tone adjustment \cite{utsav2023tone}. 
4) \textit{Question Answering:} SQuAD \cite{rajpurkar-etal-2016-squad}. 
Datasets for these four problem types are released in English, and we translated them into the other languages via the OPUS-MT model \cite{TiedemannThottingal:EAMT2020} for Spanish, German, French, and Italian, and via the M2M100 model \cite{fan2021beyond} for Korean, Chinese, and Japanese. 
5) \textit{Grammar Correction:} original-language datasets are critical here, as machine translation systems tend to correct grammatical errors \cite{luhtaru2024no}. We use Write \& Improve for English \cite{bryant2019bea}, Merlin for Italian \cite{boyd2014merlin}, ECSpell for Chinese \cite{lv2023general}, and the GitHub Typo Corpus \cite{hagiwara-mita-2020-github} for the other languages.

To calibrate the similarity threshold used in \oursp, we reserve a set of held-out tasks that include unseen problem types and languages. Specifically, we use: (i) \textit{Translation to English} and \textit{Title Generation} as novel problem types, and (ii) Portuguese, Turkish, and Serbian as novel languages.
TED Talks \cite{qi2018pretrained} serves as the dataset for translation, and XLSum \cite{hasan2021xl} for title generation. 

All datasets are split into training, validation, and test partitions; see the Appendix for details.

\begin{figure*}[t]
    \centering
        \includegraphics[width=\textwidth]{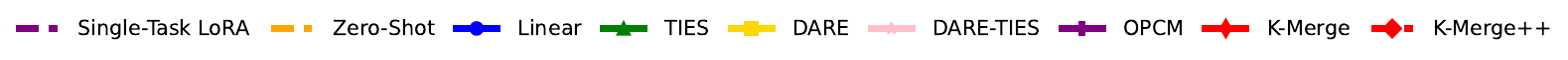}
    \begin{subfigure}[b]{0.495\textwidth}
    \centering
    \includegraphics[width=\textwidth]{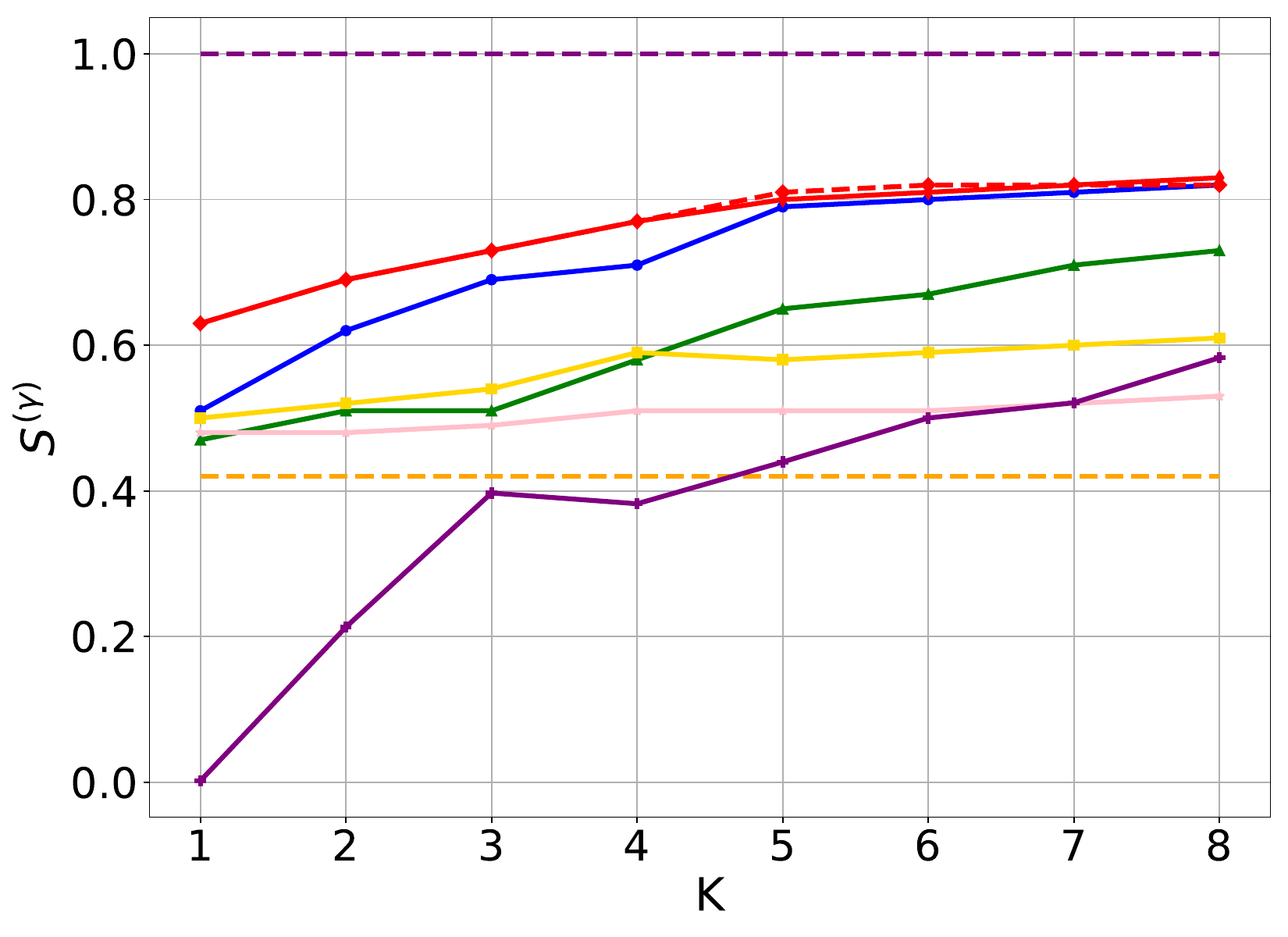}
    \caption{Llama-3.2-1B-Instruct.} \label{fig:s_over_k_lama}
    \end{subfigure}
    \begin{subfigure}[b]{0.495\textwidth}
    \centering
    \includegraphics[width=\textwidth]{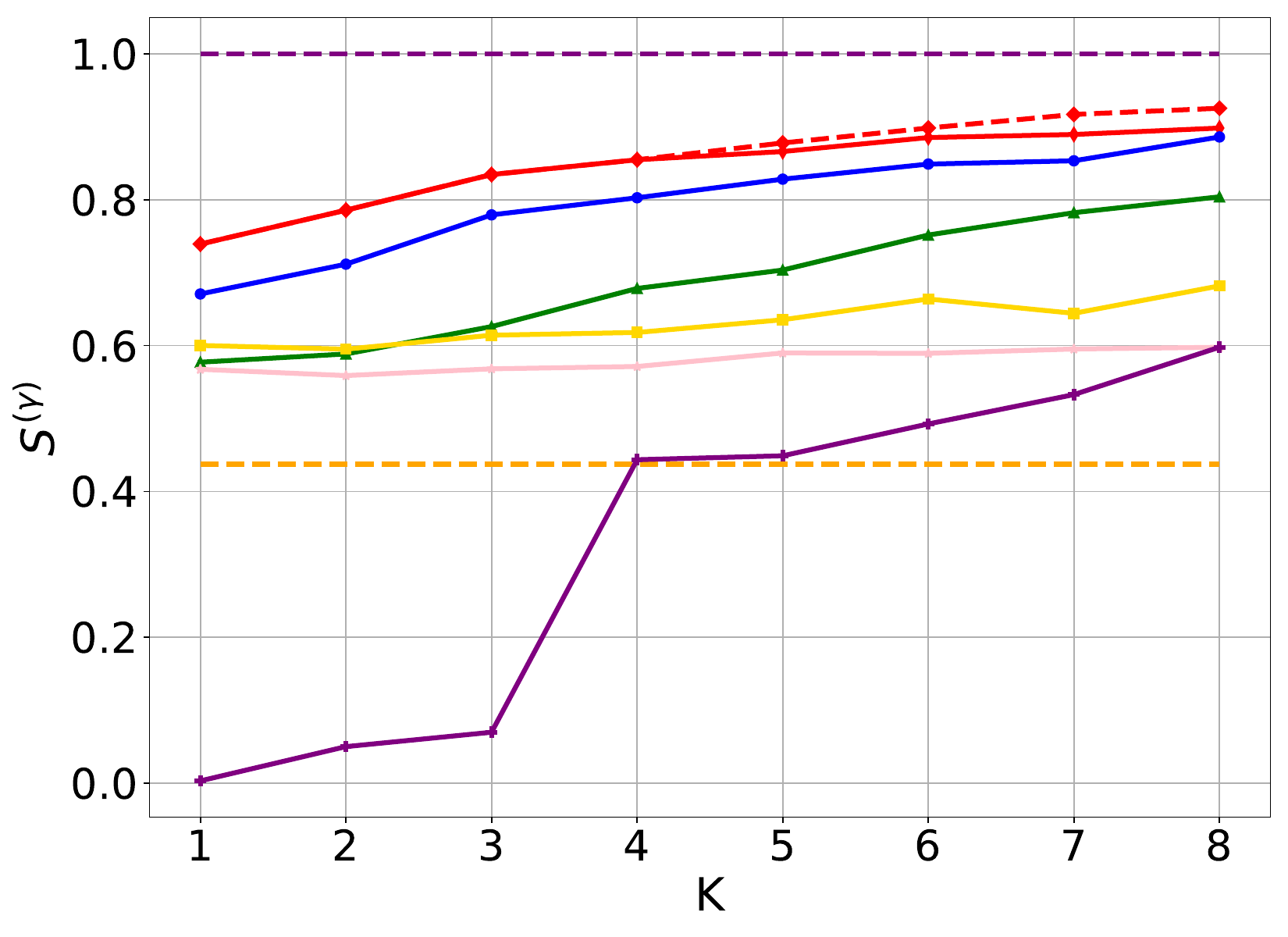}
    \caption{Qwen-2.5-1.5B-Instruct.}
    \end{subfigure}
    \caption{Score $S^{(\gamma)}$ of compared methods at variable storage budget $K$. Averaged over 3 random task orderings.}
    \label{fig:loras_CM}
\end{figure*}

\keypoint{Models} 
We experiment with two lightweight, instruction-tuned models designed for on-device deployment:
Llama-3.2-1B-Instruct 
and Qwen-2.5-1.5B-Instruct.

\keypoint{Baselines} 
We compare our solutions with several alternative approaches. 
We include zero-shot and single-task LoRA as lower and upper performance boundaries. 
We also compare to standard model merging approaches (Linear,
TIES,
DARE,
DARE-TIES) 
and continual merging (OPCM).
All merging methods are adapted to our online setting by applying them to the pair consisting of the incoming LoRA $L^{(t)}$ and the most similar on-device LoRA $L_c$.
Task descriptions are always provided via prompts (see the Appendix for examples).

\keypoint{Hyperparameters} 
All LoRAs are trained using AdamW with a learning rate of 5e-5, dropout rate of 0.05, and a mini-batch size of 3.
LoRAs use rank 32 and scaling factor $\alpha = 128$.
Linear merging uses equal weights (i.e., 0.5) for pairwise merging, while TIES, DARE, and DARE-TIES use unary weights and density 0.5.
The similarity threshold is set using the held-out LoRA set to $s = 0.020$ and $s=0.028$ for Llama-3.2-1B and Qwen-2.5-1.5B models, respectively.
In our experiments, the number of LoRAs that can be stored on-device, $K$, varies from 1 to 8. 
In total, we simulate the sequential arrival of $\gamma=40$ LoRAs. 
All results are averaged over 3 random permutations of LoRA arrival ordering (unless otherwise stated).

\subsection{Experimental Results}

\keypoint{Results with Random Task Ordering}
Fig.~\ref{fig:loras_CM} illustrates the average score of the methods for the two LLMs at varying numbers of clusters over three random task orderings. Additionally, we plot zero-shot and single-task LoRA performance as a reference. 
First, we note the great benefit of LoRA fine-tuning for all LLMs, as indicated by the improvement of single-task LoRAs over zero-shot performance. 
Second, we observe DARE, DARE-TIES, and OPCM are unable to offer competitive performance, being significantly lower than single-task LoRAs and other approaches, although the corresponding performance gap shrinks slightly as the number of clusters increases. 
TIES generally outperforms these methods; however, the Linear merge proves to be the most effective baseline. 
Remarkably, our proposed methods outperform Linear at all storage budgets. In particular, with Llama-3.2-1B our method scales better than competitors for small numbers of clusters. With Qwen-2.5-1.5B the gain is stable across different values of $K$.

Finally, {K-Merge$++$} improves the performance of {K-Merge} even further by utilizing the similarity threshold mechanism, especially for larger $K$, where there is more scope for more distinct multi-tasking LoRAs. However, the main benefit of {K-Merge$++$} is robustness against worst-case task orderings, as we analyse later.
Using our methods, the overall score reaches about 80-90\% of the single-task performance with just $8$ clusters, showing the usefulness of our solution.

\begin{table}[t]
\begin{center}
\setlength{\tabcolsep}{3.5pt}
  \resizebox{0.5\textwidth}{!}{%
    \begin{tabular}{l c|c c c}
    \toprule
    Method & Ordering & \textbf{$K=3$} & \textbf{$K=5$} & \textbf{$K=7$} \\
    \midrule
    Linear &       Random   & 0.69 & 0.79 &     0.81\\
    Linear &  Problem Types &  - & 0.80 & - \\
    Linear &  Worst &  0.62 & 0.69 & 0.73 \\ \midrule
    TIES &       Random    & 0.51 & 0.65 & 0.71 \\
    TIES &  Problem Types &  - & 0.72 & - \\
    TIES &  Worst &  0.52 & 0.56 & 0.67 \\ \midrule
    \ours  &     Random  & 0.73 & 0.80   & 0.82 \\
    \ours    & Problem Types   & - & 0.84   & - \\
    \ours   & Worst   & 0.73 & 0.75   & 0.78 \\ \midrule
    \oursp   & Random & 0.73 & 0.81   & 0.82  \\
    \oursp &  Worst  & 0.73 & 0.80   & 0.82  \\ 
    \bottomrule
    \end{tabular}
    }
\end{center}
\caption{{Robustness of approaches across different ordering of tasks evaluated by means of score $S^{(\gamma)}$ on Llama-3.2-1B.}}
\label{tab:method_vs_k}
\end{table}

\keypoint{Robustness to Task Ordering}
We investigate the robustness of our methods to different task orderings in Table~\ref{tab:method_vs_k}. 
In particular, we compare the average score obtained via random task ordering (\textit{Random}) against a clustering of LoRAs done according to problem types (\textit{Problem Types}) and an ordering that we empirically found to lead to low performance for our setup (\textit{Worst}). 
In particular, in the latter, we force LoRAs of the same problem type to appear consecutively.
We remark that we could only test \textit{Problem Types} for $K\!= \! \alpha \! = \! 5$ since we allocate one problem type for each on-device LoRA cluster.
Additionally, evaluating \oursp\ \ on the \textit{Problem Types} case is not meaningful, as the clustering is pre-determined and the similarity threshold is therefore not employed.

First, the scores for the \textit{Problem Types} case are higher than for \textit{Random}, being a favorable case where similar LoRAs are merged together (see Appendix Fig.~\ref{fig:sim_matrix}). 
Nonetheless, this ordering makes three unrealistic assumptions: 
(i) the device knows \textit{a priori} all future problem types; 
(ii) the device has exactly $K=\alpha$ on-device LoRA storage slots; and
(iii) LoRAs always have the highest similarity with other LoRAs from the same problem type rather than from different ones.
In practice, these assumptions are difficult to meet, hence motivating the usefulness of our approach. Even more, K-Merge outperforms the baselines also in this setup.

Second, under the \textit{Worst} case, we observe \oursp~ in particular significantly outperforms \ours~and shows strong robustness thanks to its threshold mechanism. Linear and TIES degrade significantly.
This setting complicates the initialization of empty slots, often causing the system to fill multiple slots with similar LoRAs (e.g., targeting the same problem type across different languages).
These results highlight the robustness of our \oursp~approach in handling \textit{a-priori} unknown and adversarial task orderings on the device side.

\begin{figure}[t]
    \centering
    \begin{subfigure}[b]{0.5\columnwidth}
    \centering
    \includegraphics[width=\linewidth]{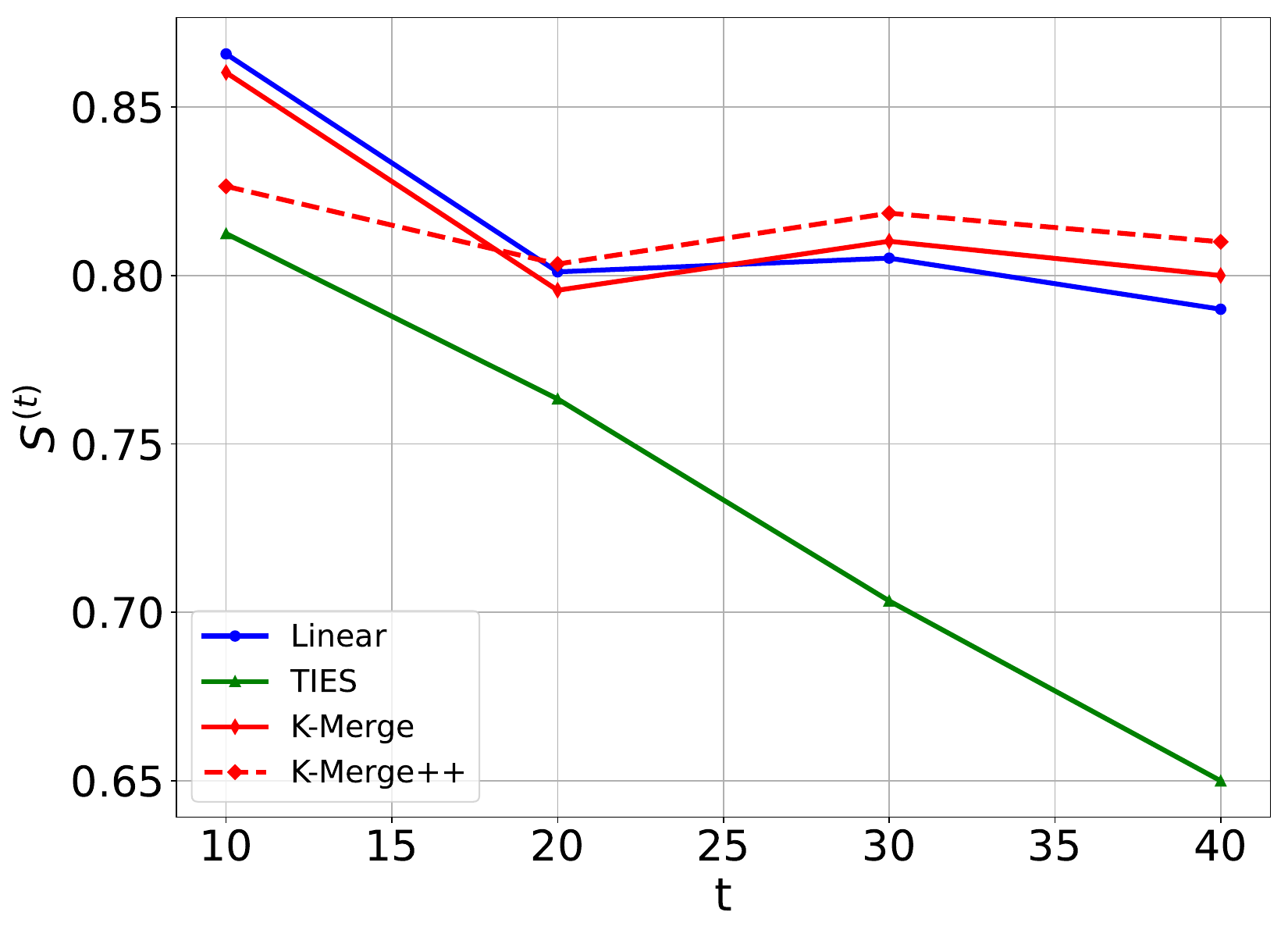}
    \end{subfigure}%
    \begin{subfigure}[b]{0.5\columnwidth}
    \centering
    \includegraphics[width=\linewidth]{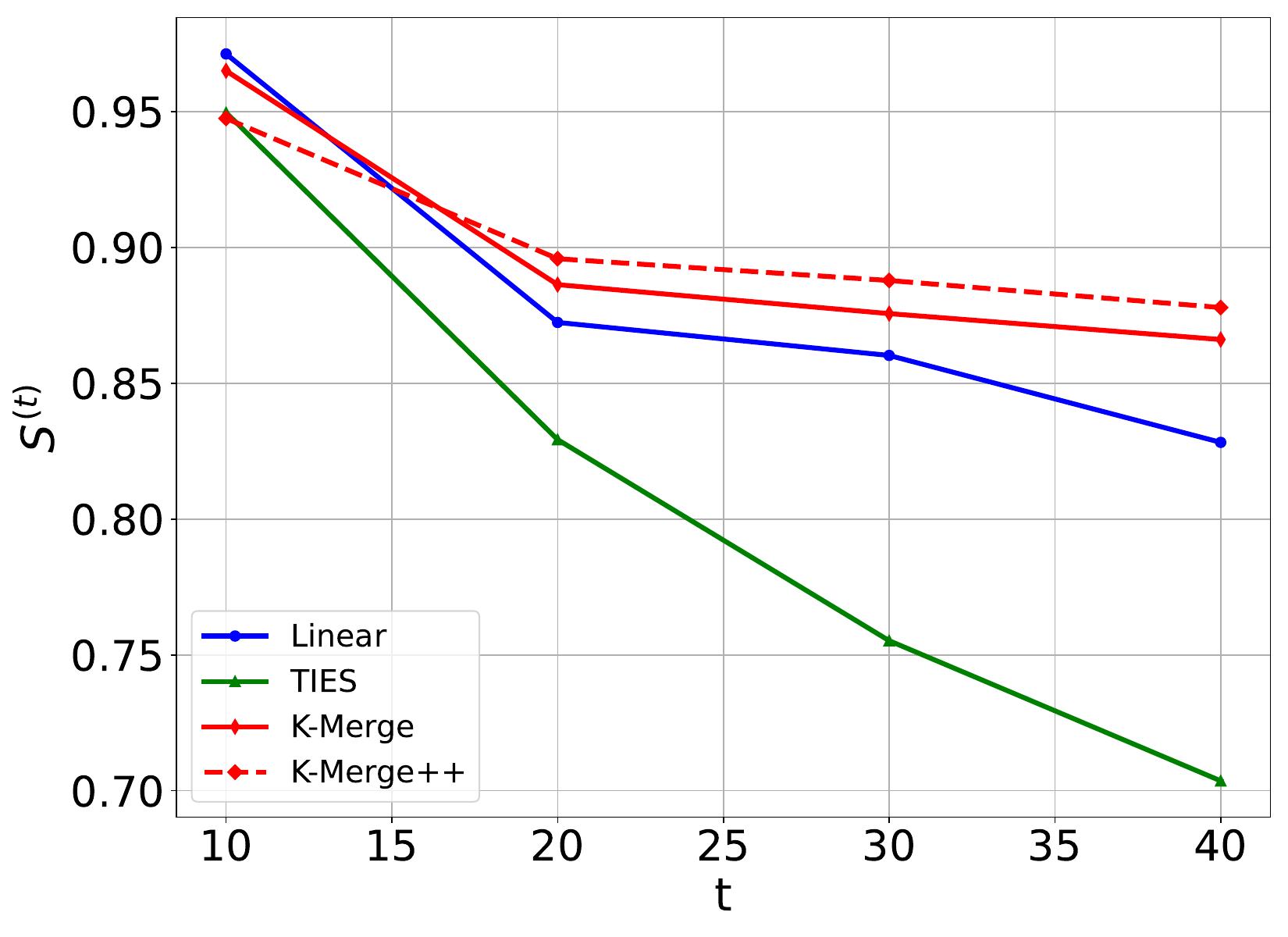}
    \end{subfigure}
    \caption{Performance $S^{(t)}$ on discovered tasks at varying timesteps for $K=5$ using Llama-3.2-1B (left) and Qwen-2.5-1.5B (right).}
    \label{fig:varying_timestep}
\end{figure}

\keypoint{Varying Timestep} Fig.~\ref{fig:varying_timestep} shows that K-Merge initially outperforms K-Merge$++$, but is quickly overtaken as the latter benefits from saving storage slots at early stages thus having them available when more diverse LoRAs arrive. Linear matches our methods at initial timesteps, but is then consistently surpassed as more adapters arrive.

\keypoint{Efficiency} 
\ours~ is lightweight and fast: the integration of a new incoming LoRA takes between 0.04s and 0.18s when there are 2 to 8 stored multi-tasking LoRAs, demonstrating its practicality for on-device deployment. \oursp~ exhibits comparable runtime. Since memory is dominated by text generation, our approach adds only negligible extra memory overhead. Full details are presented in the Appendix.

\keypoint{Storage Analysis} 
We report the number of parameters and storage used per adapter in Table~\ref{tab:num_params}. Storing, e.g., 5 adapters is lightweight, consuming 135-170MB of storage in total. Storing all 40 adapters would require about 1GB of storage, a rather substantial amount.
\begin{table}[t]
\begin{center}
\setlength{\tabcolsep}{7pt}
\resizebox{\columnwidth}{!}{
\begin{tabular}{l c c}
\toprule
Model & Parameters per LoRA & Storage per LoRA \\ \midrule
Llama-3.2-1B & 23M & 27MB \\
Qwen-2.5-1.5B & 37M & 34MB \\
\bottomrule
\end{tabular}}
\end{center}
\caption{{Number of parameters and storage used by one LoRA adapter. Storing all adapters (40) is not practical as it would take a significant amount of storage (1GB).}}
\label{tab:num_params}
\end{table}

\keypoint{Similarity Metric}
We have evaluated different similarity measures, with cosine similarity (i.e., our selected option) giving the best performance. In particular, we have compared it with the $L^1, L^2$ and $L^{\infty}$ measures. In our experiments, the cosine similarity yielded the best performances, as can be seen from Table~\ref{tab:similarity_combined}. It shows how our approach achieves $S^{(\gamma)}$ scores of 0.75 and 0.78 with $K=3$ and $K=5$, respectively. The alternative similarity measures led to significantly lower scores than cosine similarity.
\begin{table}[t]
    \centering
    \resizebox{0.85\columnwidth}{!}{
    \begin{tabular}{c c c c c}
        \toprule
        Method & $L^1$ & $L^2$ & $L^\infty$ & Ours (cos) \\
        \midrule
        \multicolumn{5}{c}{$K=3$} \\
        \midrule
        \ours   & 0.66 & 0.66 & 0.66 & 0.75 \\
        Linear  & 0.51 & 0.51 & 0.51 & 0.66 \\
        TIES    & 0.47 & 0.47 & 0.50 & 0.51 \\
        DARE    & 0.53 & 0.52 & 0.51 & 0.53 \\
        \midrule
        \multicolumn{5}{c}{$K=5$} \\
        \midrule
        \ours   & 0.69 & 0.67 & 0.67 & 0.78 \\
        Linear  & 0.50 & 0.52 & 0.57 & 0.77 \\
        TIES    & 0.52 & 0.51 & 0.52 & 0.59 \\
        DARE    & 0.52 & 0.51 & 0.51 & 0.58 \\
        \bottomrule
    \end{tabular}}
    \caption{Comparison of similarity metrics.
    Score $S^{(\gamma)}$ for different values of $K$ (single seed).
    }
    \label{tab:similarity_combined}
\end{table}

\begin{table}[t]
    \centering
    \resizebox{0.85\columnwidth}{!}{
    \begin{tabular}{c c c}
        \toprule
        Method & History-aware & Constant \\
        \midrule
        \multicolumn{3}{c}{$K=1$} \\
        \midrule
        Linear     & 0.63 & 0.45 \\
        TIES       & 0.44 & 0.46 \\
        DARE       & 0.43 & 0.50 \\
        DARE-TIES  & 0.43 & 0.48 \\
        \midrule
        \multicolumn{3}{c}{$K=3$} \\
        \midrule
        Linear     & 0.75 & 0.66 \\
        TIES       & 0.46 & 0.51 \\
        DARE       & 0.44 & 0.53 \\
        DARE-TIES  & 0.44 & 0.48 \\
        \bottomrule
    \end{tabular}}
    \caption{Effectiveness of our history-aware merging. Score $S^{(\gamma)}$ for different values of $K$ (single seed).}
    \label{tab:history_combined}
\end{table}

\keypoint{History-Aware Merging}
Note that the Linear approach combined with history-aware merging corresponds to our K-Merge. For this reason, we also apply similarity-based clustering to each merging approach that we report in the results. We have conducted additional experiments that integrate history-aware merging into the diverse baselines. The results in Table~\ref{tab:history_combined} show that they obtain worse performance than our solution (linear with history-aware merging).

\keypoint{Further Analyses} 
We include further analyses studying LoRA similarity, similarity threshold, clustering consistency and out-of-domain LoRA performance in the Appendix. The analyses show, for example, there is more similarity between LoRAs from the same problem type rather than from the same language.

\section{Conclusion}
\label{sec:conclusion}
We introduced the novel task of on-device continual LoRA merging, where incoming adapters must be merged over time due to storage constraints, with the goal of supporting new capabilities while preserving existing ones.
We proposed two methods: \ours, which merges incoming adapters with their closest stored counterpart when storage is full, and \oursp, which uses a similarity threshold to decide between merging or allocating a new slot.
Both methods have been evaluated on 40 tasks across 5 domains and 8 languages, achieving strong performance and high efficiency, while requiring no data. Average performance is similar, but \oursp\ \ is notably more robust to the task arrival order.

\section*{Limitations}
\label{sec:limit}
We restrict our study to LLMs, leaving extensions to multimodal tasks for future work. We only studied LoRA adapters as it is the current industry standard, but other adapter types could be considered. We evaluated LLMs that are in sizes suitable for on-device deployment as that is the desired use case. Our solution could also be evaluated for larger LLMs, which would, however, require significantly larger amount of compute.

\section*{Ethical Considerations}
\label{sec:ethics}

Our work focuses on text generation, which is an area where ethical considerations are important due to the potential for large-scale real-world impact. Our work in particular focuses on efficiency and as a result it does not introduce new capabilities that could be misused. However, merging of adapters can lead to weakening of safeguarding mechanisms. As a result, it is important to conduct experiments that explore how effective earlier safeguarding mechanisms remain before deploying the solution.

\bibliography{references}

\begin{thebibliography}{59}
\providecommand{\natexlab}[1]{#1}

\bibitem[{Bohdal et~al.(2026)Bohdal, Ceritli, Ozay, Moon, Lee, Ko, and Michieli}]{bohdal2025data}
Ondrej Bohdal, Taha Ceritli, Mete Ozay, Jijoong Moon, Kyeng-Hun Lee, Hyeonmok Ko, and Umberto Michieli. 2026.
\newblock \href {https://arxiv.org/abs/2601.17441} {Data-driven clustering and merging of adapters for on-device large language models}.
\newblock In \emph{ICASSP}.

\bibitem[{Bohdal et~al.(2025)Bohdal, Ozay, Moon, Lee, Ko, and Michieli}]{bohdal2025compositional}
Ondrej Bohdal, Mete Ozay, Jijoong Moon, Kyeng-Hun Lee, Hyeonmok Ko, and Umberto Michieli. 2025.
\newblock \href {https://aclanthology.org/2025.emnlp-main.1429/} {Efficient compositional multi-tasking for on-device large language models}.
\newblock In \emph{EMNLP}.

\bibitem[{Borzunov et~al.(2024)Borzunov, Ryabinin, Chumachenko, Baranchuk, Dettmers, Belkada, Samygin, and Raffel}]{borzunov2024distributed}
Alexander Borzunov, Max Ryabinin, Artem Chumachenko, Dmitry Baranchuk, Tim Dettmers, Younes Belkada, Pavel Samygin, and Colin~A Raffel. 2024.
\newblock \href {https://proceedings.neurips.cc/paper_files/paper/2023/hash/28bf1419b9a1f908c15f6195f58cb865-Abstract-Conference.html} {Distributed inference and fine-tuning of large language models over the internet}.
\newblock In \emph{NeurIPS}.

\bibitem[{Boyd et~al.(2014)Boyd, Hana, Nicolas, Meurers, Wisniewski, Abel, Sch{\"o}ne, Stindlov{\'a}, and Vettori}]{boyd2014merlin}
Adriane Boyd, Jirka Hana, Lionel Nicolas, Detmar Meurers, Katrin Wisniewski, Andrea Abel, Karin Sch{\"o}ne, Barbora Stindlov{\'a}, and Chiara Vettori. 2014.
\newblock \href {https://aclanthology.org/L14-1488/} {The merlin corpus: Learner language and the cefr.}
\newblock In \emph{LREC}.

\bibitem[{Bryant et~al.(2019)Bryant, Felice, Andersen, and Briscoe}]{bryant2019bea}
Christopher Bryant, Mariano Felice, {\O}istein~E Andersen, and Ted Briscoe. 2019.
\newblock \href {https://aclanthology.org/W19-4406/} {{The BEA-2019 shared task on grammatical error correction}}.
\newblock In \emph{ACL Workshop}.

\bibitem[{Burke(2023)}]{burke2023aicore}
Dave Burke. 2023.
\newblock \href {https://android-developers.googleblog.com/2023/12/a-new-foundation-for-ai-on-android.html} {A new foundation for ai on android}.

\bibitem[{Ceritli et~al.(2025)Ceritli, Bohdal, Ozay, Moon, Lee, Ko, and Michieli}]{ceritli2025hydraopt}
Taha Ceritli, Ondrej Bohdal, Mete Ozay, Jijoong Moon, Kyenghun Lee, Hyeonmok Ko, and Umberto Michieli. 2025.
\newblock \href {https://aclanthology.org/2025.emnlp-main.1365/} {Hydraopt: Navigating the efficiency-performance trade-off of adapter merging}.
\newblock In \emph{EMNLP}.

\bibitem[{Coleman et~al.(2024)Coleman, Quarantiello, Hurtado, and Lomonaco}]{colemanadaptive}
Eric~Nuertey Coleman, Luigi Quarantiello, Julio Hurtado, and Vincenzo Lomonaco. 2024.
\newblock \href {https://openreview.net/forum?id=tlB5eonGEk} {Adaptive lora merging for efficient domain incremental learning}.
\newblock In \emph{NeurIPS Workshop}.

\bibitem[{Dhar et~al.(2021)Dhar, Guo, Liu, Tripathi, Kurup, and Shah}]{dhar2021ondevice}
Sauptik Dhar, Junyao Guo, Jiayi~(Jason) Liu, Samarth Tripathi, Unmesh Kurup, and Mohak Shah. 2021.
\newblock \href {https://dl.acm.org/doi/10.1145/3450494} {A survey of on-device machine learning: An algorithms and learning theory perspective}.
\newblock \emph{ACM TIOT}.

\bibitem[{Ding et~al.(2022)Ding, Qin, Yang, Wei, Yang, Su, Hu, Chen, Chan, Chen et~al.}]{ding2022delta}
Ning Ding, Yujia Qin, Guang Yang, Fuchao Wei, Zonghan Yang, Yusheng Su, Shengding Hu, Yulin Chen, Chi-Min Chan, Weize Chen, and 1 others. 2022.
\newblock \href {https://arxiv.org/abs/2203.06904} {Delta tuning: A comprehensive study of parameter efficient methods for pre-trained language models}.
\newblock \emph{arXiv preprint arXiv:2203.06904}.

\bibitem[{Dubey et~al.(2024)Dubey, Jauhri, Pandey, Kadian, Al-Dahle, Letman, Mathur, Schelten, Yang, Fan et~al.}]{dubey2024llama}
Abhimanyu Dubey, Abhinav Jauhri, Abhinav Pandey, Abhishek Kadian, Ahmad Al-Dahle, Aiesha Letman, Akhil Mathur, Alan Schelten, Amy Yang, Angela Fan, and 1 others. 2024.
\newblock \href {https://arxiv.org/abs/2407.21783} {The llama 3 herd of models}.
\newblock \emph{arXiv preprint arXiv:2407.21783}.

\bibitem[{Dziadzio et~al.(2025)Dziadzio, Udandarao, Roth, Prabhu, Akata, Albanie, and Bethge}]{dziadzio2024merge}
Sebastian Dziadzio, Vishaal Udandarao, Karsten Roth, Ameya Prabhu, Zeynep Akata, Samuel Albanie, and Matthias Bethge. 2025.
\newblock \href {https://openaccess.thecvf.com/content/CVPR2025/html/Dziadzio_How_to_Merge_Your_Multimodal_Models_Over_Time_CVPR_2025_paper.html} {How to merge your multimodal models over time?}
\newblock In \emph{CVPR}.

\bibitem[{Einolghozati et~al.(2020)Einolghozati, Gupta, Diedrick, and Gupta}]{einolghozati2020sound}
Arash Einolghozati, Anchit Gupta, Keith Diedrick, and Sonal Gupta. 2020.
\newblock \href {https://aclanthology.org/2020.emnlp-main.414/} {Sound natural: Content rephrasing in dialog systems}.
\newblock In \emph{EMNLP}.

\bibitem[{Fan et~al.(2021)Fan, Bhosale, Schwenk, Ma, El-Kishky, Goyal, Baines, Celebi, Wenzek, Chaudhary et~al.}]{fan2021beyond}
Angela Fan, Shruti Bhosale, Holger Schwenk, Zhiyi Ma, Ahmed El-Kishky, Siddharth Goyal, Mandeep Baines, Onur Celebi, Guillaume Wenzek, Vishrav Chaudhary, and 1 others. 2021.
\newblock \href {https://dl.acm.org/doi/abs/10.5555/3546258.3546365} {Beyond english-centric multilingual machine translation}.
\newblock \emph{JMLR}.

\bibitem[{Gauthier-Caron et~al.(2024)Gauthier-Caron, Siriwardhana, Stein, Ehghaghi, Goddard, McQuade, Solawetz, and Labonne}]{gauthier2024merging}
Thomas Gauthier-Caron, Shamane Siriwardhana, Elliot Stein, Malikeh Ehghaghi, Charles Goddard, Mark McQuade, Jacob Solawetz, and Maxime Labonne. 2024.
\newblock \href {https://arxiv.org/abs/2410.08371} {Merging in a bottle: Differentiable adaptive merging (dam) and the path from averaging to automation}.
\newblock \emph{arXiv preprint arXiv:2410.08371}.

\bibitem[{Gliwa et~al.(2019)Gliwa, Mochol, Biesek, and Wawer}]{gliwa2019samsum}
Bogdan Gliwa, Iwona Mochol, Maciej Biesek, and Aleksander Wawer. 2019.
\newblock \href {https://aclanthology.org/D19-5409/} {Samsum corpus: A human-annotated dialogue dataset for abstractive summarization}.
\newblock In \emph{EMNLP-IJCNLP Workshop}.

\bibitem[{Gunter et~al.(2024)Gunter, Wang, Wang, Pang, Narayanan, Zhang, Zhang, Chen, Chiu, Qiu et~al.}]{gunter2024apple}
Tom Gunter, Zirui Wang, Chong Wang, Ruoming Pang, Andy Narayanan, Aonan Zhang, Bowen Zhang, Chen Chen, Chung-Cheng Chiu, David Qiu, and 1 others. 2024.
\newblock \href {https://arxiv.org/abs/2407.21075} {Apple intelligence foundation language models}.
\newblock \emph{arXiv preprint arXiv:2407.21075}.

\bibitem[{Hagiwara and Mita(2020)}]{hagiwara-mita-2020-github}
Masato Hagiwara and Masato Mita. 2020.
\newblock \href {https://aclanthology.org/2020.lrec-1.835/} {{G}it{H}ub typo corpus: A large-scale multilingual dataset of misspellings and grammatical errors}.
\newblock In \emph{LREC}.

\bibitem[{Hammoud et~al.(2024)Hammoud, Michieli, Pizzati, Torr, Bibi, Ghanem, and Ozay}]{hammoud2024model}
Hasan Abed Al~Kader Hammoud, Umberto Michieli, Fabio Pizzati, Philip Torr, Adel Bibi, Bernard Ghanem, and Mete Ozay. 2024.
\newblock \href {https://aclanthology.org/2024.findings-emnlp.762/} {Model merging and safety alignment: One bad model spoils the bunch}.
\newblock In \emph{EMNLP Findings}.

\bibitem[{Han et~al.(2024)Han, Gao, Liu, Zhang, and Zhang}]{han2024parameter}
Zeyu Han, Chao Gao, Jinyang Liu, Jeff Zhang, and Sai~Qian Zhang. 2024.
\newblock \href {https://openreview.net/forum?id=lIsCS8b6zj} {Parameter-efficient fine-tuning for large models: A comprehensive survey}.
\newblock In \emph{TMLR}.

\bibitem[{Hasan et~al.(2021)Hasan, Bhattacharjee, Islam, Mubasshir, Li, Kang, Rahman, and Shahriyar}]{hasan2021xl}
Tahmid Hasan, Abhik Bhattacharjee, Md.~Saiful Islam, Kazi Mubasshir, Yuan-Fang Li, Yong-Bin Kang, M.~Sohel Rahman, and Rifat Shahriyar. 2021.
\newblock \href {https://aclanthology.org/2021.findings-acl.413/} {{XL}-sum: Large-scale multilingual abstractive summarization for 44 languages}.
\newblock In \emph{ACL-IJCNLP Findings}.

\bibitem[{Hu et~al.(2022)Hu, Shen, Wallis, Allen-Zhu, Li, Wang, Wang, and Chen}]{hu2021lora}
Edward~J Hu, Yelong Shen, Phillip Wallis, Zeyuan Allen-Zhu, Yuanzhi Li, Shean Wang, Lu~Wang, and Weizhu Chen. 2022.
\newblock \href {https://openreview.net/forum?id=nZeVKeeFYf9} {Lora: Low-rank adaptation of large language models}.
\newblock In \emph{ICLR}.

\bibitem[{Huang et~al.(2024)Huang, Liu, Lin, Pang, Du, and Lin}]{huang2023lorahub}
Chengsong Huang, Qian Liu, Bill~Yuchen Lin, Tianyu Pang, Chao Du, and Min Lin. 2024.
\newblock \href {https://openreview.net/forum?id=TrloAXEJ2B} {Lorahub: Efficient cross-task generalization via dynamic lora composition}.
\newblock In \emph{COLM}.

\bibitem[{Ilharco et~al.(2023)Ilharco, Ribeiro, Wortsman, Gururangan, Schmidt, Hajishirzi, and Farhadi}]{ilharco2022editing}
Gabriel Ilharco, Marco~Tulio Ribeiro, Mitchell Wortsman, Suchin Gururangan, Ludwig Schmidt, Hannaneh Hajishirzi, and Ali Farhadi. 2023.
\newblock \href {https://openreview.net/forum?id=6t0Kwf8-jrj} {Editing models with task arithmetic}.
\newblock In \emph{ICLR}.

\bibitem[{Jandaghi et~al.(2024)Jandaghi, Sheng, Bai, Pujara, and Sidahmed}]{jandaghi2023faithful}
Pegah Jandaghi, Xianghai Sheng, Xinyi Bai, Jay Pujara, and Hakim Sidahmed. 2024.
\newblock \href {https://aclanthology.org/2024.findings-acl.904/} {Faithful persona-based conversational dataset generation with large language models}.
\newblock In \emph{ACL Findings}.

\bibitem[{Kopiczko et~al.(2024)Kopiczko, Blankevoort, and Asano}]{kopiczko2023vera}
Dawid~J Kopiczko, Tijmen Blankevoort, and Yuki~M Asano. 2024.
\newblock \href {https://openreview.net/forum?id=NjNfLdxr3A} {{VeRA: Vector-based random matrix adaptation}}.
\newblock In \emph{ICLR}.

\bibitem[{Liu et~al.(2024{\natexlab{a}})Liu, Wang, Yin, Molchanov, Wang, Cheng, and Chen}]{liu2024dora}
Shih-Yang Liu, Chien-Yi Wang, Hongxu Yin, Pavlo Molchanov, Yu-Chiang~Frank Wang, Kwang-Ting Cheng, and Min-Hung Chen. 2024{\natexlab{a}}.
\newblock \href {https://dl.acm.org/doi/10.5555/3692070.3693369} {Dora: Weight-decomposed low-rank adaptation}.
\newblock In \emph{ICML}.

\bibitem[{Liu et~al.(2024{\natexlab{b}})Liu, Shi, He, Ye, Fabbri, Liu, Radev, and Cohan}]{liu2023learning}
Yixin Liu, Kejian Shi, Katherine~S He, Longtian Ye, Alexander~R Fabbri, Pengfei Liu, Dragomir Radev, and Arman Cohan. 2024{\natexlab{b}}.
\newblock \href {https://aclanthology.org/2024.naacl-long.478/} {On learning to summarize with large language models as references}.
\newblock In \emph{NAACL}.

\bibitem[{Luhtaru et~al.(2024)Luhtaru, Korotkova, and Fishel}]{luhtaru2024no}
Agnes Luhtaru, Elizaveta Korotkova, and Mark Fishel. 2024.
\newblock \href {https://aclanthology.org/2024.eacl-long.73/} {No error left behind: Multilingual grammatical error correction with pre-trained translation models}.
\newblock In \emph{EACL}.

\bibitem[{Lv et~al.(2023)Lv, Cao, Geng, Ai, Yan, and Fu}]{lv2023general}
Qi~Lv, Ziqiang Cao, Lei Geng, Chunhui Ai, Xu~Yan, and Guohong Fu. 2023.
\newblock \href {https://dl.acm.org/doi/10.1145/3564271} {General and domain-adaptive chinese spelling check with error-consistent pretraining}.
\newblock \emph{ACM TALLIP}.

\bibitem[{Mao et~al.(2025)Mao, Ge, Fan, Xu, Mi, Hu, and Gao}]{mao2025survey}
Yuren Mao, Yuhang Ge, Yijiang Fan, Wenyi Xu, Yu~Mi, Zhonghao Hu, and Yunjun Gao. 2025.
\newblock \href {https://link.springer.com/article/10.1007/s11704-024-40663-9} {A survey on lora of large language models}.
\newblock \emph{Frontiers of Computer Science}.

\bibitem[{Marczak et~al.(2024)Marczak, Twardowski, Trzci{\'n}ski, and Cygert}]{marczak2024magmax}
Daniel Marczak, Bart{\l}omiej Twardowski, Tomasz Trzci{\'n}ski, and Sebastian Cygert. 2024.
\newblock \href {https://www.ecva.net/papers/eccv_2024/papers_ECCV/html/11489_ECCV_2024_paper.php} {Magmax: Leveraging model merging for seamless continual learning}.
\newblock In \emph{ECCV}.

\bibitem[{Minaee et~al.(2024)Minaee, Mikolov, Nikzad, Chenaghlu, Socher, Amatriain, and Gao}]{minaee2024large}
Shervin Minaee, Tomas Mikolov, Narjes Nikzad, Meysam Chenaghlu, Richard Socher, Xavier Amatriain, and Jianfeng Gao. 2024.
\newblock \href {https://arxiv.org/abs/2402.06196} {Large language models: A survey}.
\newblock \emph{arXiv preprint arXiv:2402.06196}.

\bibitem[{Qi et~al.(2018)Qi, Sachan, Felix, Padmanabhan, and Neubig}]{qi2018pretrained}
Ye~Qi, Devendra Sachan, Matthieu Felix, Sarguna Padmanabhan, and Graham Neubig. 2018.
\newblock \href {https://aclanthology.org/N18-2084/} {When and why are pre-trained word embeddings useful for neural machine translation?}
\newblock In \emph{NAACL}.

\bibitem[{{Qwen Team}(2024)}]{qwen2.5}
{Qwen Team}. 2024.
\newblock \href {https://qwenlm.github.io/blog/qwen2.5/} {Qwen2.5: A party of foundation models}.

\bibitem[{Rajpurkar et~al.(2016)Rajpurkar, Zhang, Lopyrev, and Liang}]{rajpurkar-etal-2016-squad}
Pranav Rajpurkar, Jian Zhang, Konstantin Lopyrev, and Percy Liang. 2016.
\newblock \href {https://aclanthology.org/D16-1264/} {{SQ}u{AD}: 100,000+ questions for machine comprehension of text}.
\newblock In \emph{EMNLP}.

\bibitem[{Renduchintala et~al.(2024)Renduchintala, Konuk, and Kuchaiev}]{renduchintala2024tied}
Adithya Renduchintala, Tugrul Konuk, and Oleksii Kuchaiev. 2024.
\newblock \href {https://aclanthology.org/2024.naacl-long.481/} {Tied-{L}o{RA}: Enhancing parameter efficiency of {L}o{RA} with weight tying}.
\newblock In \emph{NAACL}.

\bibitem[{Rothe et~al.(2021)Rothe, Mallinson, Malmi, Krause, and Severyn}]{severyn2021grammar}
Sascha Rothe, Jonathan Mallinson, Eric Malmi, Sebastian Krause, and Aliaksei Severyn. 2021.
\newblock \href {https://aclanthology.org/2021.acl-short.89/} {A simple recipe for multilingual grammatical error correction}.
\newblock In \emph{ACL}.

\bibitem[{Shenaj et~al.(2025)Shenaj, Bohdal, Ozay, Zanuttigh, and Michieli}]{shenaj2024lora}
Donald Shenaj, Ondrej Bohdal, Mete Ozay, Pietro Zanuttigh, and Umberto Michieli. 2025.
\newblock \href {https://openaccess.thecvf.com/content/ICCV2025/html/Shenaj_LoRA.rar_Learning_to_Merge_LoRAs_via_Hypernetworks_for_Subject-Style_Conditioned_ICCV_2025_paper.html} {Lora.rar: Learning to merge loras via hypernetworks for subject-style conditioned image generation}.
\newblock In \emph{ICCV}.

\bibitem[{Shu et~al.(2024)Shu, Luo, Hoskere, Zhu, Liu, Tong, Chen, and Meng}]{shu2024rewritelm}
Lei Shu, Liangchen Luo, Jayakumar Hoskere, Yun Zhu, Yinxiao Liu, Simon Tong, Jindong Chen, and Lei Meng. 2024.
\newblock \href {https://dl.acm.org/doi/10.1609/aaai.v38i17.29863} {Rewritelm: An instruction-tuned large language model for text rewriting}.
\newblock In \emph{AAAI}.

\bibitem[{Sokar et~al.(2025)Sokar, Dziugaite, Arnab, Iscen, Castro, and Schmid}]{sokar2025continual}
Ghada Sokar, Gintare~Karolina Dziugaite, Anurag Arnab, Ahmet Iscen, Pablo~Samuel Castro, and Cordelia Schmid. 2025.
\newblock \href {https://arxiv.org/abs/2506.03189} {Continual learning in vision-language models via aligned model merging}.
\newblock \emph{arXiv preprint arXiv:2506.03189}.

\bibitem[{Sticha et~al.(2024)Sticha, Braunschweiler, Doddipatla, and Knill}]{sticha2024qa}
Abigail Sticha, Norbert Braunschweiler, Rama~Sanand Doddipatla, and Kate~M Knill. 2024.
\newblock \href {https://dl.acm.org/doi/abs/10.1145/3640794.3665573} {Advancing faithfulness of large language models in goal-oriented dialogue question answering}.
\newblock In \emph{ACM CUI}.

\bibitem[{Stoica et~al.(2025)Stoica, Ramesh, Ecsedi, Choshen, and Hoffman}]{stoica2025model}
George Stoica, Pratik Ramesh, Boglarka Ecsedi, Leshem Choshen, and Judy Hoffman. 2025.
\newblock \href {https://openreview.net/forum?id=67X93aZHII} {Model merging with {SVD} to tie the knots}.
\newblock In \emph{ICLR}.

\bibitem[{Tang et~al.(2025)Tang, Yang, Shen, Luo, Hu, Zhang, Du, and Tao}]{tang2025merging}
Anke Tang, Enneng Yang, Li~Shen, Yong Luo, Han Hu, Lefei Zhang, Bo~Du, and Dacheng Tao. 2025.
\newblock \href {https://openreview.net/forum?id=rdGMyTPhui} {Merging on the fly without retraining: A sequential approach to scalable continual model merging}.
\newblock In \emph{NeurIPS}.

\bibitem[{Tiedemann and Thottingal(2020)}]{TiedemannThottingal:EAMT2020}
J{\"o}rg Tiedemann and Santhosh Thottingal. 2020.
\newblock \href {https://aclanthology.org/2020.eamt-1.61/} {{OPUS-MT} — {B}uilding open translation services for the {W}orld}.
\newblock In \emph{EAMT}.

\bibitem[{Utsav(2023)}]{utsav2023tone}
Kumar Utsav. 2023.
\newblock Redpajama-incite-base-3b-v1 model finetuned for paraphrasing and changing the tone.
\newblock \url{https://huggingface.co/llm-toys}.

\bibitem[{Wortsman et~al.(2022)Wortsman, Ilharco, Gadre, Roelofs, Gontijo-Lopes, Morcos, Namkoong, Farhadi, Carmon, Kornblith et~al.}]{wortsman2022model}
Mitchell Wortsman, Gabriel Ilharco, Samir~Ya Gadre, Rebecca Roelofs, Raphael Gontijo-Lopes, Ari~S Morcos, Hongseok Namkoong, Ali Farhadi, Yair Carmon, Simon Kornblith, and 1 others. 2022.
\newblock \href {https://proceedings.mlr.press/v162/wortsman22a.html} {Model soups: averaging weights of multiple fine-tuned models improves accuracy without increasing inference time}.
\newblock In \emph{ICML}.

\bibitem[{Xiao et~al.(2024)Xiao, Liu, Zhang, and Xing}]{xiao2024lm}
Shitao Xiao, Zheng Liu, Peitian Zhang, and Xingrun Xing. 2024.
\newblock \href {https://aclanthology.org/2024.findings-acl.145/} {{LM-Cocktail}: Resilient tuning of language models via model merging}.
\newblock In \emph{ACL Findings}.

\bibitem[{Yadav et~al.(2023)Yadav, Tam, Choshen, Raffel, and Bansal}]{yadav2024ties}
Prateek Yadav, Derek Tam, Leshem Choshen, Colin~A Raffel, and Mohit Bansal. 2023.
\newblock \href {https://proceedings.neurips.cc/paper_files/paper/2023/hash/1644c9af28ab7916874f6fd6228a9bcf-Abstract-Conference.html} {Ties-merging: Resolving interference when merging models}.
\newblock In \emph{NeurIPS}.

\bibitem[{Yang et~al.(2024{\natexlab{a}})Yang, Yang, Hui, Zheng, Yu, Zhou, Li, Li, Liu, Huang, Dong, Wei, Lin, Tang, Wang, Yang, Tu, Zhang, Ma, Xu, Zhou, Bai, He, Lin, Dang, Lu, Chen, Yang, Li, Xue, Ni, Zhang, Wang, Peng, Men, Gao, Lin, Wang, Bai, Tan, Zhu, Li, Liu, Ge, Deng, Zhou, Ren, Zhang, Wei, Ren, Fan, Yao, Zhang, Wan, Chu, Liu, Cui, Zhang, and Fan}]{qwen2}
An~Yang, Baosong Yang, Binyuan Hui, Bo~Zheng, Bowen Yu, Chang Zhou, Chengpeng Li, Chengyuan Li, Dayiheng Liu, Fei Huang, Guanting Dong, Haoran Wei, Huan Lin, Jialong Tang, Jialin Wang, Jian Yang, Jianhong Tu, Jianwei Zhang, Jianxin Ma, and 40 others. 2024{\natexlab{a}}.
\newblock \href {https://arxiv.org/abs/2407.10671} {Qwen2 technical report}.
\newblock \emph{arXiv preprint arXiv:2407.10671}.

\bibitem[{Yang et~al.(2024{\natexlab{b}})Yang, Shen, Guo, Wang, Cao, Zhang, and Tao}]{yang2024model}
Enneng Yang, Li~Shen, Guibing Guo, Xingwei Wang, Xiaochun Cao, Jie Zhang, and Dacheng Tao. 2024{\natexlab{b}}.
\newblock \href {https://arxiv.org/abs/2408.07666} {Model merging in llms, mllms, and beyond: Methods, theories, applications and opportunities}.
\newblock \emph{arXiv preprint arXiv:2408.07666}.

\bibitem[{Yu et~al.(2024)Yu, Yu, Yu, Huang, and Li}]{yu2024language}
Le~Yu, Bowen Yu, Haiyang Yu, Fei Huang, and Yongbin Li. 2024.
\newblock \href {https://openreview.net/forum?id=fq0NaiU8Ex} {Language models are super mario: Absorbing abilities from homologous models as a free lunch}.
\newblock In \emph{ICML}.

\bibitem[{Zhang et~al.(2021)Zhang, Wang, Deb, Zheng, Shokouhi, and Awadallah}]{zhang2021dataset}
Mozhi Zhang, Wei Wang, Budhaditya Deb, Guoqing Zheng, Milad Shokouhi, and Ahmed~Hassan Awadallah. 2021.
\newblock \href {https://aclanthology.org/2021.acl-long.97/} {A dataset and baselines for multilingual reply suggestion}.
\newblock In \emph{ACL}.

\bibitem[{Zhang et~al.(2023)Zhang, Chen, Bukharin, Karampatziakis, He, Cheng, Chen, and Zhao}]{zhang2023adalora}
Qingru Zhang, Minshuo Chen, Alexander Bukharin, Nikos Karampatziakis, Pengcheng He, Yu~Cheng, Weizhu Chen, and Tuo Zhao. 2023.
\newblock \href {https://openreview.net/forum?id=lq62uWRJjiY} {Adaptive budget allocation for parameter-efficient fine-tuning}.
\newblock In \emph{ICLR}.

\bibitem[{Zhao et~al.(2023)Zhao, Zhou, Li, Tang, Wang, Hou, Min, Zhang, Zhang, Dong et~al.}]{zhao2023survey}
Wayne~Xin Zhao, Kun Zhou, Junyi Li, Tianyi Tang, Xiaolei Wang, Yupeng Hou, Yingqian Min, Beichen Zhang, Junjie Zhang, Zican Dong, and 1 others. 2023.
\newblock \href {https://arxiv.org/abs/2303.18223} {A survey of large language models}.
\newblock \emph{arXiv preprint arXiv:2303.18223}.

\bibitem[{Zhao et~al.(2025)Zhao, Shen, Zhu, Li, Su, Wang, and Wu}]{zhao2025merging}
Ziyu Zhao, Tao Shen, Didi Zhu, Zexi Li, Jing Su, Xuwu Wang, and Fei Wu. 2025.
\newblock \href {https://openreview.net/forum?id=j6fsbpAllN} {Merging {LoRA}s like playing {LEGO}: Pushing the modularity of lo{RA} to extremes through rank-wise clustering}.
\newblock In \emph{ICLR}.

\bibitem[{Zhu et~al.(2024{\natexlab{a}})Zhu, Liu, Dong, Xu, Huang, Kong, Chen, and Li}]{zhu2023multilingual}
Wenhao Zhu, Hongyi Liu, Qingxiu Dong, Jingjing Xu, Shujian Huang, Lingpeng Kong, Jiajun Chen, and Lei Li. 2024{\natexlab{a}}.
\newblock \href {https://aclanthology.org/2024.findings-naacl.176/} {Multilingual machine translation with large language models: Empirical results and analysis}.
\newblock In \emph{NAACL Findings}.

\bibitem[{Zhu et~al.(2024{\natexlab{b}})Zhu, Li, Liu, Ma, and Wang}]{zhu2024survey}
Xunyu Zhu, Jian Li, Yong Liu, Can Ma, and Weiping Wang. 2024{\natexlab{b}}.
\newblock \href {https://aclanthology.org/2024.tacl-1.85/} {A survey on model compression for large language models}.
\newblock \emph{TACL}.

\bibitem[{Zi et~al.(2023)Zi, Qi, Wang, Wang, Wong, and Zhang}]{zi2023delta}
Bojia Zi, Xianbiao Qi, Lingzhi Wang, Jianan Wang, Kam-Fai Wong, and Lei Zhang. 2023.
\newblock \href {https://arxiv.org/abs/2309.02411} {Delta-lora: Fine-tuning high-rank parameters with the delta of low-rank matrices}.
\newblock \emph{arXiv preprint arXiv:2309.02411}.

\end{thebibliography}

\appendix

\clearpage

\appendix
\noindent {\Large\bfseries Appendix \par} 
\vspace{0.5\baselineskip}

\renewcommand{\thesection}{A.\arabic{section}}
\renewcommand{\thefigure}{A.\arabic{figure}}
\renewcommand{\thetable}{A.\arabic{table}}
\renewcommand{\thealgorithm}{A.\arabic{algorithm}}

\setcounter{equation}{0}
\setcounter{figure}{0}
\setcounter{table}{0}
\setcounter{section}{0}

\noindent This document provides additional material that could not be included in the main paper due to space constraints.
Additional details for the employed datasets, prompts, and hardware are reported in Appendix~\ref{app:sec:additional_details}.
Detailed results for single-task LoRAs and continual merging are reported in Appendix~\ref{app:sec:detailed_results}.
Additional analyses describing computation time overhead of our approach, cosine similarity histograms, clustering consistency and out-of-domain performance are presented in Appendix~\ref{app:sec:analyses}.

\section{Additional Details}
\label{app:sec:additional_details}

\subsection{Dataset Statistics}
Table~\ref{tab:datasets} shows the number of training, validation, and test samples for the various problem types and languages.

\subsection{Prompts}
Table~\ref{tab:prompts_full} shows the prompts that have been used for the different problem types, which are taken from \cite{ceritli2025hydraopt}. Different tones have separate prompts, which are also detailed in Table~\ref{tab:prompts_full}.

\subsection{Hardware and Software Details}
\label{sec:A_HW}
For all our experiments, we have used Python 3.9.21 with PyTorch 2.6.0+cu124.
We have run our experiments on an Ubuntu 20.04.6 LTS machine with kernel version 5.4.0-205-generic, equipped with 10 NVIDIA A40 GPUs, and two Intel Xeon Gold 5218 processors (16 cores per socket). Note that the continual merging approach (i.e., our contribution) is very lightweight and fast, the hardware resources have been used mostly to train the LoRAs and perform inference.

\section{Detailed Results}
\label{app:sec:detailed_results}

\begin{table}[!h]
    \centering
  \resizebox{0.5\textwidth}{!}{%
    \begin{tabular}{c c r r r r r r r}
        \toprule
         & & \multicolumn{5}{c}{Task Performance} \\
       Lang & Method & TC & QA & RP & Sum & Tone & AVG & S\\ 
        \midrule
        
        \multirow{6}{*}{en}  & &  \multicolumn{5}{c}{Llama-3.2-1B-Instruct} \\
         & Zero-Shot & 13.1 & 16.2 & 5.1 & 23.4 & 27.6 & 17.1 & 0.39 \\ 
         & LoRA & 35.1 & 63.5 & 23.0 & 38.2 & 58.1 & 43.6 & 1\\ 
         \cmidrule{2-9}
        & &  \multicolumn{5}{c}{Qwen 2.5 1.5B} \\
          & Zero-shot & 19.6	& 24.6	& 5.0	& 27.5	& 43.5	& 24.0	& 0.51\\
         & LoRA & 38.6 &	68.0 &	22.3 &	38.6 &	59.0	& 45.3 & 1  \\
        \hline
        
        \multirow{6}{*}{de} & &  \multicolumn{5}{c}{Llama-3.2-1B-Instruct} \\
        & Zero-Shot & 7.6 & 8.4 & 2.2 & 15.8 & 17.2 & 10.3 & 0.33 \\
         & LoRA & 26.0 & 36.5 & 10.7 & 28.6 & 43.8 & 29.1 & 1\\
                  \cmidrule{2-9}
        & &  \multicolumn{5}{c}{Qwen 2.5 1.5B} \\
         & Zero-shot & 7.2	&10.1	&2.3	&18.7	&33.9	&14.4	&0.43\\
        & LoRA &23.0	&42.4	&12.8	&28.9	&44.0&	30.2 & 1 \\
        \hline

        \multirow{6}{*}{es}& &  \multicolumn{5}{c}{Llama-3.2-1B-Instruct} \\
         & Zero-Shot & 6.7 & 9.1 & 3.5 & 16.8 & 25.4 & 12.3 & 0.36\\
         & LoRA & 34.0 & 42.3 & 13.2 & 31.3 & 44.4 & 33.0 & 1\\
                  \cmidrule{2-9}
        & &  \multicolumn{5}{c}{Qwen 2.5 1.5B} \\
         & Zero-shot & 12.1 &	13.0 &	2.7 &	21.2 &	38.5 &	17.5 &	0.47  \\
         & LoRA & 29.7 &	48.7 &	14.7 &	32.6 &	46.0 &	34.3 & 1  \\
        \hline
        
        \multirow{6}{*}{fr} & &  \multicolumn{5}{c}{Llama-3.2-1B-Instruct} \\
         & Zero-Shot & 5.7 & 7.6 & 3.9 & 18.0 & 26.6 & 12.4 & 0.40\\
         & LoRA & 20.5 & 34.7 & 12.2 & 30.2 & 46.2 & 28.8 & 1 \\
                  \cmidrule{2-9}
        & &  \multicolumn{5}{c}{Qwen 2.5 1.5B} \\
         & Zero-shot & 10.4 &	12.3	& 3.1	& 20.8	& 30.3	& 15.4 &	0.42  \\
         & LoRA & 36.8 &	38.7 &	14.3	& 31.3	& 47.5 &	33.7 & 1\\
        \hline
        
        \multirow{6}{*}{it} & &  \multicolumn{5}{c}{Llama-3.2-1B-Instruct} \\
         & Zero-Shot & 23.3 & 7.2 & 2.5 & 17.8 & 21.3 & 14.4 & 0.49 \\
         & LoRA & 27.2 & 37.1 & 8.8 & 28.3 & 42.8 & 28.8 & 1\\
                  \cmidrule{2-9}
        & &  \multicolumn{5}{c}{Qwen 2.5 1.5B} \\
        & Zero-shot & 22.9	& 9.9	& 1.8	& 18.7	& 32.2	& 17.1	& 0.49 \\
        & LoRA & 31.6 &	42.0 &	11.6 &	30.2 &	43.7 &	31.8  & 1 \\
                \hline
        
        \multirow{6}{*}{ja}& &  \multicolumn{5}{c}{Llama-3.2-1B-Instruct} \\
         & Zero-Shot & 3.7 & 6.8 & 4.3 & 18.4 & 34.6 & 13.6 & 0.52\\
         & LoRA & 12.9 & 22.5 & 9.1 & 27.5 & 39.4 & 22.3 & 1\\
                  \cmidrule{2-9}
        & &  \multicolumn{5}{c}{Qwen 2.5 1.5B} \\
        & Zero-shot & 5.2 &	7.1	& 3.2	& 18.7 &	32.8& 	13.4	& 0.45 \\
        & LoRA & 26.4 &	27.9 &	11.6 &	26.7 &	39.4 &	26.4  & 1 \\
        \hline
                
        \multirow{6}{*}{ko} & &  \multicolumn{5}{c}{Llama-3.2-1B-Instruct} \\
         & Zero-Shot & 2.0 & 3.2 & 1.2 & 8.1 & 23.2 & 7.6 & 0.40 \\
         & LoRA & 8.2 & 18.2 & 5.0 & 14.0 & 30.5 & 15.2 & 1\\
                  \cmidrule{2-9}
        & &  \multicolumn{5}{c}{Qwen 2.5 1.5B} \\
        & Zero-shot &3.3 &	4.5 &	0.5 &	8.1 &	16.8 &	6.7 &	0.30 \\
        & LoRA & 21.5 &	23.3 &	6.2	 & 15.2 &	31.5 &	19.5  & 1  \\
                \hline

        \multirow{6}{*}{zh} & &  \multicolumn{5}{c}{Llama-3.2-1B-Instruct} \\
         & Zero-Shot & 3.0 & 7.5 & 2.0 & 18.1 & 25.8 & 11.3 & 0.44\\
         & LoRA & 37.9 & 22.6 & 7.5 & 22.7 & 35.2 & 25.2 & 1\\
                  \cmidrule{2-9}
        & &  \multicolumn{5}{c}{Qwen 2.5 1.5B} \\
        & Zero-shot & 11.9 &	7.4 &	1.9 &	18.3 &	27.4 &	13.4 &	0.43 \\
        & LoRA & 63.2 &	29.1 &	6.6 &	26.1 &	37.3 &	32.5 & 1\\

        \bottomrule
    \end{tabular}
    }
       \caption{{Test performance on the main tasks (\%). F-0.5 score for Text Correction (TC), F-1 for Question Answering (QA), Weighted ROUGE for Smart Reply (RP) and ROUGE-L for Text Summarization (Sum) and Tone Adjustment (Tone). }
       }
    \label{tab:lora_domain_performance_samsum}
\end{table}

\subsection{Single-task LoRA Performance}
Table~\ref{tab:lora_domain_performance_samsum} shows the performance of zero-shot (i.e., without any LoRA adapter) inference and the ones of the single-task LoRA for all combinations of problem types and languages. 
It is possible to notice  a large gap between the two options as using LoRAs typically doubles the score. 
Qwen-2.5-1.5B has better performances than Llama-3.2-1B, possibly due to the larger number of parameters. 
The scores vary across languages, with English outperforming the others.
We provide a similar analysis in Table~\ref{tab:lora_held_out_performance} for the additional languages and problem types used in the held-out set. Note that this data has been used only to select the similarity threshold $s$ in the \oursp{} version of the approach.

\begin{table*}[t]
\begin{center}
\resizebox{\textwidth}{!}{%
\begin{tabular}{lccccc} %
\toprule
Problem Type & Dataset & Language & \# Train. Samples & \# Val. Samples & \# Test Samples \\
\midrule
\multirow{8}{*}{\begin{tabular}{l}Grammar Error\\
                                  Correction
                \end{tabular}} & Write \& Improve & English & 23,523 & 2,526 & 2,639 \\
& Merlin & Italian & 572 & 79 & 81\\
& ECSpell & Chinese & 6,680 & 750 & 750\\
& GitHub Typo Corpus & French & 616 & 240 & 227\\
& GitHub Typo Corpus & German & 412 & 119 & 132\\
& GitHub Typo Corpus & Japanese & 1,043 & 325 & 321\\
& GitHub Typo Corpus & Korean & 255 & 75 & 93\\
& GitHub Typo Corpus & Spanish & 348 & 137 & 116\\
\midrule
Smart Reply & Persona-Chat Synthetic & All & 225,061 & 1,000 & 1,000 \\
Text Summarization & SAMSum & All & 14,732 & 818 & 819 \\
Tone Adj. & Sound Natural & All & 2,245 & 321 & 642 \\
Question Answering & SQuAD & All & 65,699 & 1,000 & 1,000 \\
\bottomrule
\end{tabular}%
}
\end{center}
\caption{Summary of the statistics for the employed datasets.}
\label{tab:datasets}
\end{table*}
\begin{table}[t]
    \centering
  \resizebox{0.45\textwidth}{!}{%
    \begin{tabular}{c c c c}
        \toprule
       Lang & Method & Translation & Title generation \\ 
        \midrule
        \multirow{2}{*}{pt} & Zero-Shot & 21.6 & 19.4 \\ 
         & LoRA & 41.4 & 21.4 \\ 
        \hline
        \multirow{2}{*}{tr} & Zero-Shot & 13.2 & 19.9 \\ 
         & LoRA & 17.4 & 21.2 \\ 
        \hline
        \multirow{2}{*}{sr} & Zero-Shot & 9.7 & 12.1 \\ 
         & LoRA & 27.2 & 10.8 \\ 
        \bottomrule
    \end{tabular}
    }
       \caption{Test performance on the held-out tasks (\%). Languages: Portuguese, Turkish, and Serbian. Problem types: translation (to English) and Title generation. BLEU score is used for translation and ROUGE-L for title generation.
       }
    \label{tab:lora_held_out_performance}
\end{table}
\subsection{Detailed Continual Merging Results}
We analyzed our continual merging experiments in Fig.~\ref{fig:loras_CM} in the main text, while in this section, we report the corresponding results in more detail, including the mean and standard deviation across three random task orderings in Table~\ref{tab:llama} and Table~\ref{tab:qwen}.
The first table shows the results for the Llama-3.2-1B model, while the second shows them for the Qwen-2.5-1.5B model.
\begin{table*}[t]
    \centering
    \resizebox{\textwidth}{!}{%
    \begin{tabular}{l c c c c c c c c}
         \toprule
          \multirow{2}{*}{Method} & \multicolumn{8}{c}{$K$} \\
          & 1 & 2 & 3 & 4 & 5 & 6 & 7 & 8\\
         \hline 
        Linear & $0.51\pm0.06$ &	$0.62\pm0.06$ &	$0.69\pm0.03$ &	$0.71\pm0.03$ &	$0.79\pm0.02$ &	$0.80\pm0.04$ &	$0.81\pm0.00$ &	$0.82\pm0.01$\\
        TIES & $0.47\pm0.05$ &	$0.51\pm0.03$ &	$0.51\pm0.02$ &	$0.58\pm0.00$ &	$0.65\pm0.06$ &	$0.67\pm0.01$ &	$0.71\pm0.02$ &	$0.73\pm0.02$\\
        DARE & $0.50\pm0.03$ &	$0.52\pm0.02$ &	$0.54\pm0.03$ &	$0.59\pm0.03$ &	$0.58\pm0.04$ &	$0.59\pm0.03$ &	$0.60\pm0.02$ &	$0.61\pm0.03$\\
        DARE-TIES  & $0.48\pm0.02$ &	$0.48\pm0.01$ &	$0.49\pm0.01$ &	$0.51\pm0.02$ &	$0.51\pm0.02$ &	$0.51\pm0.02$ &	$0.52\pm0.01$ &	$0.53\pm0.01$\\
        OPCM &  $0.00 \pm 0.00$& 	$0.21\pm0.14$& $0.40\pm0.04$	& $0.38\pm0.06$	& $0.44\pm0.06$	& $0.50\pm0.06$	& $0.52\pm0.03$	& $0.58\pm0.03$  \\
        K-Merge & $\mathbf{0.63\pm0.00}$ &	$\mathbf{0.69\pm0.02}$ &	$\mathbf{0.73\pm0.01}$ &	$\mathbf{0.77\pm0.00}$ &	$0.80\pm0.02$ &	$0.81\pm0.02$ &	$\mathbf{0.82\pm0.01}$ &	$\mathbf{0.83\pm0.02}$\\
        K-Merge++ & $\mathbf{0.63\pm0.00}$ &	$\mathbf{0.69\pm0.02}$ &	$\mathbf{0.73\pm0.01}$ &	$\mathbf{0.77\pm0.00}$ &	$\mathbf{0.81\pm0.02}$ &	$\mathbf{0.82\pm0.03}$ &	$\mathbf{0.82\pm0.03}$ &	$0.82\pm0.03$\\
         \bottomrule
    \end{tabular}}
    \caption{Llama-3.2-1B-Instruct: score $S^{(\gamma)}$ of compared merging methods at variable LoRA storage budget $K$. Average and standard deviation across three random task orderings. Best in bold.}
    \label{tab:llama}
\end{table*}
\begin{table*}[!h]
    \centering
    \resizebox{\textwidth}{!}{%
    \begin{tabular}{l c c c c c c c c}
         \toprule
          \multirow{2}{*}{Method} & \multicolumn{8}{c}{$K$} \\
          & 1 & 2 & 3 & 4 & 5 & 6 & 7 & 8\\
         \hline 
        Linear & $0.67\pm0.01$ &	$0.71\pm0.01$ &	$0.78\pm0.05$ &	$0.80\pm0.05$ &	$0.83\pm0.05$ &	$0.85\pm0.06$ &	$0.85\pm0.06$ &	$0.89\pm0.03$ \\
        TIES & $0.58\pm0.01$&	$0.59\pm0.01$ &	$0.63\pm0.01$	& $0.68\pm0.02$ &	$0.70\pm0.01$	& $0.75\pm0.02$	& $0.78\pm0.04$	& $0.80\pm0.02$ \\
        DARE & $0.60\pm0.01$ &	$0.60\pm0.01$ &	$0.61\pm0.02$	& $0.62\pm0.01$	& $0.64\pm0.02$	& $0.66\pm0.01$ &	$0.64\pm0.05$& 	$0.68\pm0.03$\\
        DARE-TIES & $0.57\pm0.01$ &	$0.56\pm0.00$ &	$0.57\pm0.00$	& $0.57\pm0.01$	& $0.59\pm0.01$ &	$0.59\pm0.01$ &	$0.60\pm0.01$ &	$0.60\pm0.01$\\
        OPCM &  $0.00\pm0.00$ &	$0.05\pm0.03$	& $0.07\pm0.01$ &	$0.44\pm0.13$ &	$0.45\pm0.15$ &	$0.49\pm0.21$	& $0.53\pm0.17$ & $0.60\pm0.13$ \\
        K-Merge & $\mathbf{0.74\pm0.00}$ &	$\mathbf{0.79\pm0.01}$	&$\mathbf{0.83\pm0.01}$ &	$\mathbf{0.85\pm0.01}$ &	$0.87\pm0.02$ &	$0.89\pm0.02$ &	$0.89\pm0.02$ &	$0.90\pm0.02$\\
        K-Merge++ & $\mathbf{0.74\pm0.00}$	& $\mathbf{0.79\pm0.01}$ &	$\mathbf{0.83\pm0.01}$ &	$\mathbf{0.85\pm0.01}$	& $\mathbf{0.88\pm0.01}$ &	$\mathbf{0.90\pm0.01}$ &	$\mathbf{0.92\pm0.02}$ &	$\mathbf{0.93\pm0.01}$ \\
         \bottomrule
    \end{tabular}}
    \caption{Qwen-2.5-1.5B-Instruct: score $S^{(\gamma)}$ of compared merging methods at variable LoRA storage budget $K$. Average and standard deviation across three random task orderings. Best in bold.}
    \label{tab:qwen}
\end{table*}

\section{Additional Analyses}
\label{app:sec:analyses}

\subsection{Integration Time of Incoming LoRAs}
\label{sec:A_clustering}
\begin{figure}[t]
    \centering
    \includegraphics[width=0.9\linewidth]{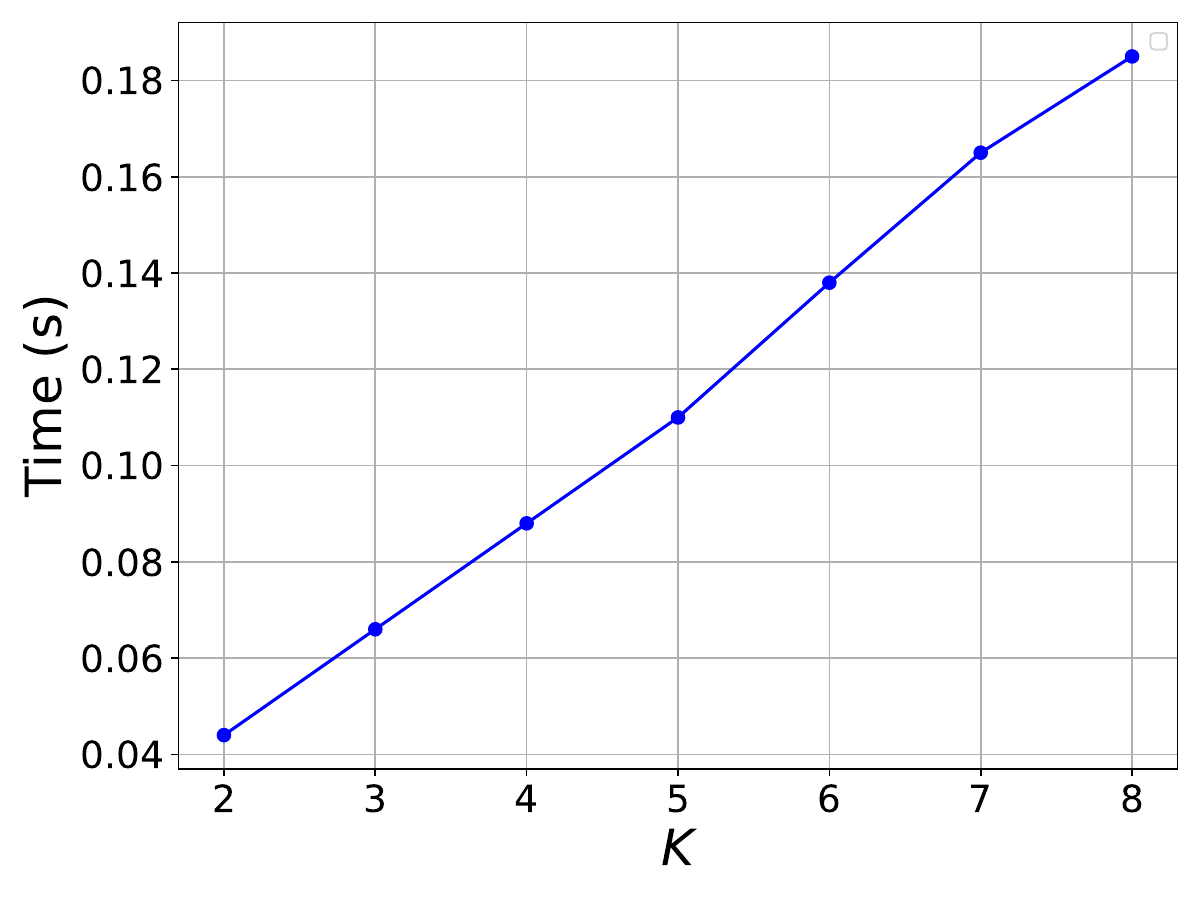}
    \caption{Analysis of the time it takes to integrate a new LoRA for the \ours{} approach (values for \oursp{} are roughly the same). The time is proportional to the number of stored LoRAs (clusters).}
    \label{fig:clustering_time}
\end{figure}

Fig.~\ref{fig:clustering_time} shows the time that it takes to integrate a new incoming LoRA using \ours{}. The time is a function of the number of stored LoRAs (clusters), and it scales roughly linearly with the number of clusters. The proposed approach is lightweight, as the time varies between 0.04 and 0.18 seconds. The time required by \oursp{} is similar to \ours{} because all additional operations take negligible time. Memory usage is dominated by text generation.

\subsection{Cosine Similarities Analysis}
\begin{figure}[!h]
    \centering
    \begin{subfigure}[b]{0.4\textwidth}
    \centering
    \includegraphics[width=\textwidth]{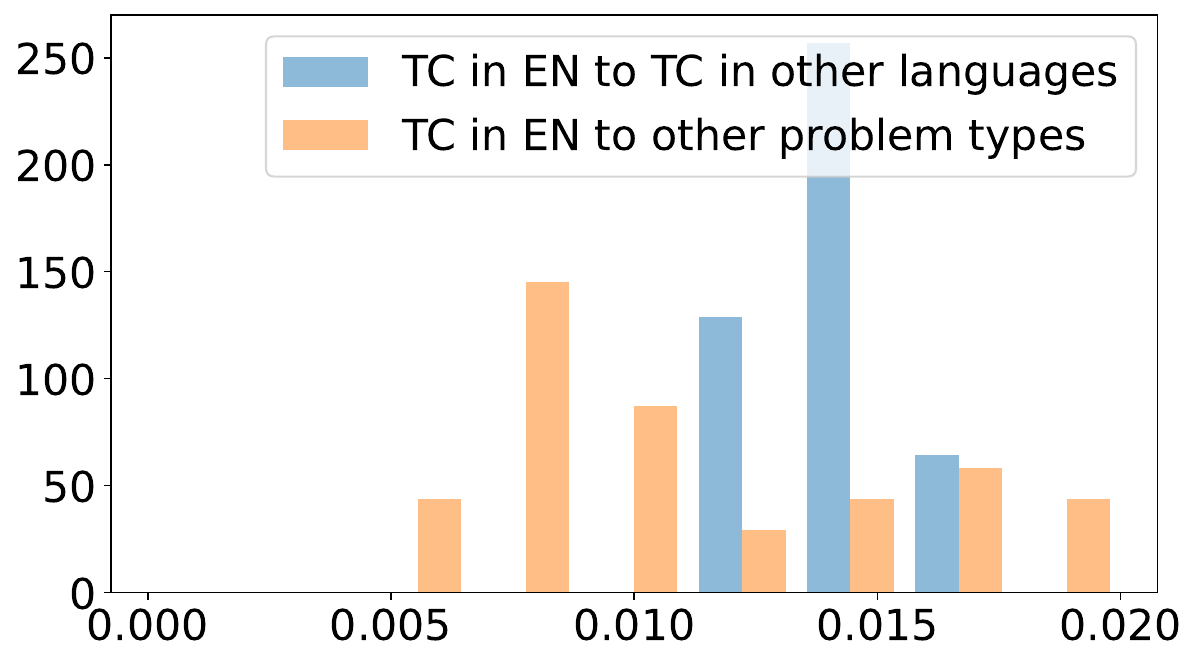}
    \caption{Text correction}
    \end{subfigure}
    \begin{subfigure}[b]{0.4\textwidth}
    \centering
    \includegraphics[width=\textwidth]{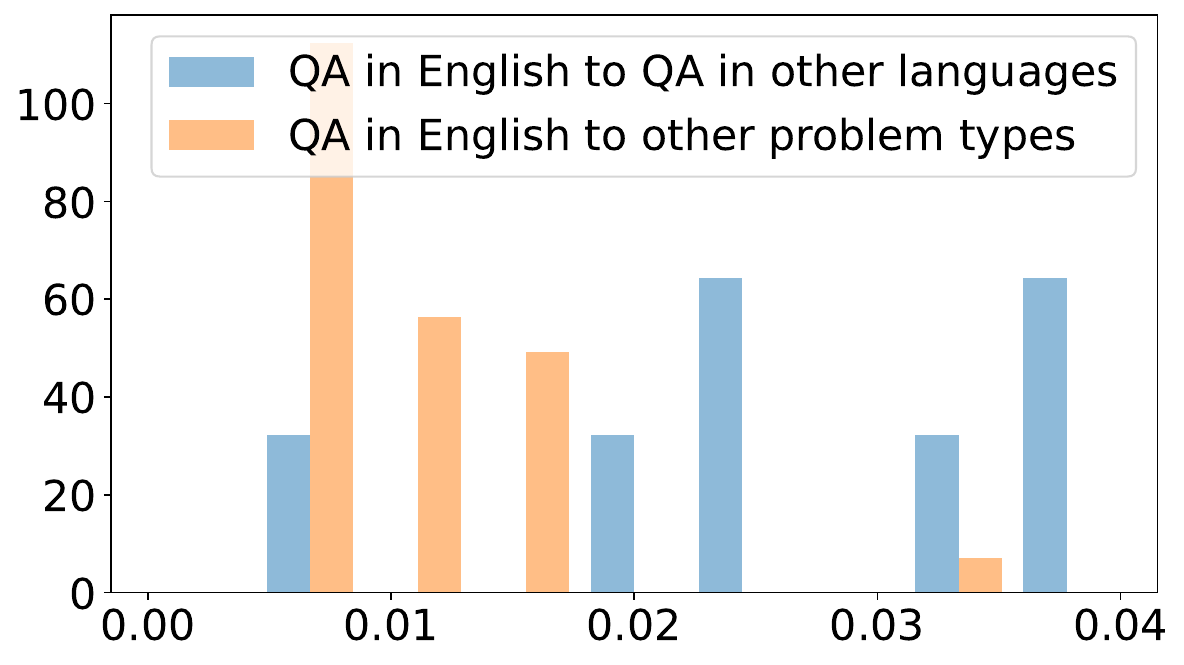}
    \caption{Question answering}
    \end{subfigure}
    \caption{LoRA cosine similarities histograms for different cases with Llama-3.2-1B. In general, there is more similarity for the same problem type in different languages rather than between problem types in the same language.}
    \label{fig:sim_histograms}
\end{figure}
\begin{figure}[!h]
    \centering
    \includegraphics[width=\linewidth]{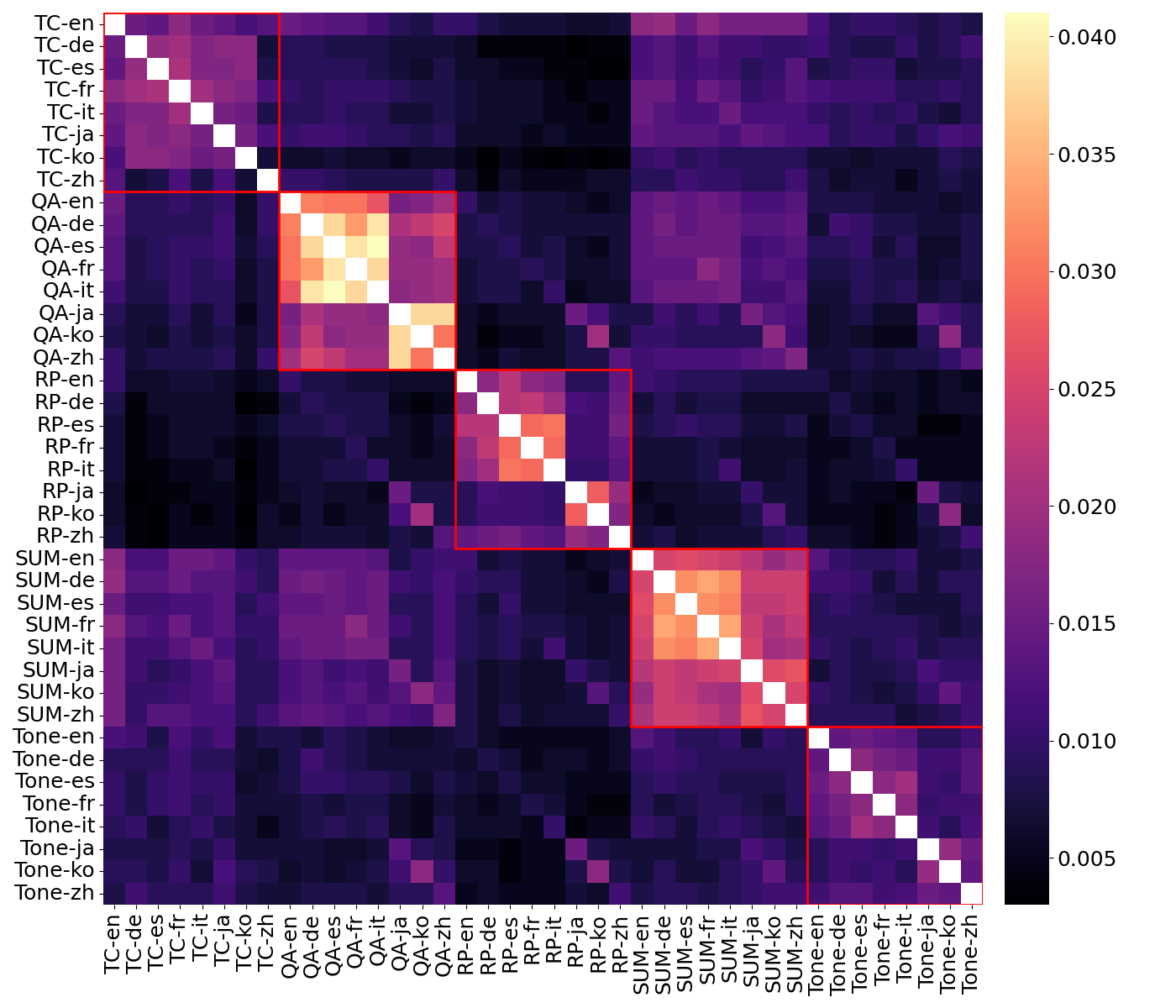}
    \caption{Similarity $\mathrm{sim}(L^{(i)},L^{(j)}), \forall i,j=1,\ldots,\gamma, i\neq j$ between LoRAs trained on Llama-3.2-1B. LoRAs tend to cluster per problem type (red solid boxes) naturally.}
    \label{fig:sim_matrix}
\end{figure}

Fig.~\ref{fig:sim_histograms} shows the histograms of LoRA cosine similarities for text correction (TC) and question answering (QA) with respect to 1) the other problem types in the same language, 2) the same problem type in different languages. We use the Llama-3.2-1B model and English as the reference language for this analysis. The results indicate that there is less similarity between different problem types in the same language than for the same problem type in different languages.
Fig.~\ref{fig:sim_matrix} reports the cross-task LoRA similarity, $\mathrm{sim}(L^{(i)},L^{(j)}), \forall i,j=1,\ldots,\gamma, i\neq j$, across both tasks and languages. 
We observe that LoRAs tend to naturally cluster per problem type (red solid boxes), while similarity per language is weaker. Nonetheless, European languages display higher similarity between themselves rather than to Asian ones, as expected.
We remark that the system is unaware of the overall task sequence at any time and, therefore, could not perform clustering based on this information.

\subsection{Clustering Consistency (at $K\!=\!\alpha$)}
To further confirm that LoRAs tend to cluster per problem type, we take the history of merges and analyse
the frequency of problem types within each cluster.
Specifically, for each cluster, we identify the most common problem type and compute the fraction of items in that cluster that belong to that problem type.
The overall clustering consistency is then calculated as the total number of items matching their cluster’s dominant problem type, divided by the total number of items.
We restrict this analysis to the case where the number of on-device LoRA slots equals the number of problem types, i.e., $K=\alpha=5$.

Linear has the lowest clustering consistency (82.5\%). \ours\ \ has a higher score (83.3\%) which is further improved by \oursp\ \ (88.3\%, which corresponds to the highest result).
Interestingly, TIES has a high clustering consistency (86.7\%); however, it underperforms other approaches.

\subsection{Similarity Threshold}
Table~\ref{tab:ablation_s} shows the performance for different values of the threshold $s$. We selected $s=0.02$ as it is the median value of pairwise similarities across LoRAs on held-out tasks, but it also performs the best compared to other options.
\begin{table}[!h]
\begin{center}
\setlength{\tabcolsep}{7pt}
\resizebox{0.45\textwidth}{!}{
\begin{tabular}{l c c c c}
\toprule
$\quad \quad s$ & 0.010 & 0.015 & 0.020 & 0.025\\ \midrule
\oursp & 0.68 & 0.74 & 0.81 & 0.80 \\
\bottomrule
\end{tabular}}
\end{center}
\caption{{Ablation of threshold $s$ on Llama-3.2-1B for $K=5$. Value of $s=0.020$, as also selected via our median rule, performs the best.}}
\label{tab:ablation_s}
\end{table}

\subsection{Larger Models}
In Table~\ref{tab:llama_3b}, we report the performance of the compared methods using Llama-3.2-3B model (instruction tuned) as the base model, confirming that our approach outperforms the competing approaches also when employing larger models.
\begin{table}[h]
    \centering
    \begin{tabular}{c c c c c}
        \toprule
         Method  & $k=1$ & $k=3$ & $k=5$ & $k=7$  \\
        \midrule
         \ours & 0.62 & 0.71     & 0.77 & 0.82 \\
         Linear  & 0.53 & 0.65     & 0.77 & 0.81 \\
         TIES    & 0.48 & 0.53     & 0.61 & 0.72 \\
         DARE    & 0.52 & 0.54 & 0.58 & 0.59 \\
        \bottomrule
    \end{tabular}
    \caption{Llama-3.2-3B-Instruct: score $S^{(\gamma)}$ of compared merging methods at variable LoRA storage budget $K$ (single seed).}
    \label{tab:llama_3b}
\end{table}

\subsection{Out-of-Domain Performance}
\begin{figure}[!h]
    \centering
    \includegraphics[width=\linewidth]{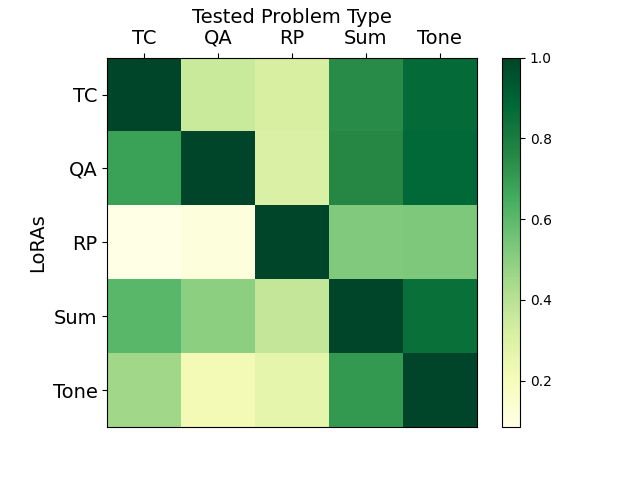}
    \caption{Generalization performance of each English LoRA to the other problem types, normalized.}
    \label{fig:lora_generalization}
\end{figure}

We analyse out-of-domain generalization performance of each English LoRA to other problem types in Fig.~\ref{fig:lora_generalization}. The scores are normalized with respect to the ones of using the corresponding problem-type LoRA (i.e., the problem-type LoRA performance corresponds to a score of 1.0). Problem type pairs such as tone adjustment and text correction, or summarization and question answering, share more similarity and have relatively high scores, while, for example, reply and tone adjustment, or reply and text correction have low scores.

\begin{CJK}{UTF8}{mj}
\begin{table*}[t]
\begin{center}
\resizebox{0.98\textwidth}{!}{%
\begin{tabular}{lcc}
\toprule
Problem Type & Language & Prompt \\
\midrule
\multirow{8}{*}{Grammar Error Correction} & English & Remove all grammatical errors from this text: \\
& Spanish & Quita todos los errores gramaticales de este texto: \\
& French & Supprimez tous les erreurs grammaticales de ce texte: \\
& German & Verbessere alle grammatischen Fehler in diesem Text: \\
& Italian & Rimuovi tutti gli errori grammaticali da questo testo: \\
& Chinese & 删除该文本中的所有语法错误: \\
& Korean & 주어진 사용자의 입력에 오타나 문법 오류가 있으면 고친다: \\
& Japanese & このテキストからすべての文法エラーを削除する: \\
\midrule
\multirow{8}{*}{Smart Reply} & English & Suggest a reply for the following text: \\
& Spanish & Sugiera una respuesta para el texto siguiente: \\
& French & Propose une réponse pour le texte suivant: \\
& German & Schlagen Sie eine Antwort für den folgenden Text vor: \\
& Italian & Suggerisci una risposta per il seguente testo: \\
& Chinese & 建议对以下文本进行回复: \\
& Korean & 다음 텍스트에 대한 답변을 제안하시오: \\
& Japanese & 次のテキストに対する返信を提案します: \\
\midrule
\multirow{8}{*}{Text Summarization} & English & Summarize the following text: \\
& Spanish & Resume el siguiente texto: \\
& French & Résume le texte suivant: \\
& German & Zusammenfassen Sie den folgenden Text: \\
& Italian & Riassumi il seguente testo: \\
& Chinese & 总结一下下面的文字: \\
& Korean & 다음 텍스트를 요약하시오: \\
& Japanese & 次の文章を要約します: \\
\midrule
\multirow{8}{*}{Tone Adj. (Professional)} & English & Changes a given user's input sentence or text to the Professional style: \\
& Spanish & Cambia la oración o el texto introducido por un usuario al estilo Profesional: \\
& French & Transforme la phrase ou le texte saisi par un utilisateur en style Professionnel: \\
& German & ändert die Eingabe eines bestimmten Benutzers in einen Professionellen Stil: \\
& Italian & Cambia la frase o il testo immesso da un utente in stile Professionale: \\
& Chinese & 将给定用户的输入句子或文本更改为专业风格: \\
& Korean & 주어진 사용자의 입력을 전문적인 문체로 변경한다: \\
& Japanese & 指定されたユーザーの入力文またはテキストを プロフェッショナル スタイルに変更する: \\
\midrule
\multirow{8}{*}{Tone Adj. (Casual)} & English & Changes a given user's input sentence or text to the Casual style: \\
& Spanish & Cambia la oración o el texto introducido por un usuario al estilo Informal: \\
& French & Transforme la phrase ou le texte saisi par un utilisateur en style Informel: \\
& German & ändert die Eingabe eines bestimmten Benutzers in einen Freundlichen Stil: \\
& Italian & Cambia la frase o il testo immesso da un utente in stile Informal: \\
& Chinese & 将给定用户的输入句子或文本更改为日常风格: \\
& Korean & 주어진 사용자의 입력을 평범한 문체로 변경한다: \\
& Japanese & 指定されたユーザーの入力文またはテキストを カジュアル スタイルに変更する: \\
\midrule
\multirow{8}{*}{Tone Adj. (Witty)} & English & Changes a given user's input sentence or text to the Witty style: \\
& Spanish & Cambia la oración o el texto introducido por un usuario al estilo Ingenioso: \\
& French & Transforme la phrase ou le texte saisi par un utilisateur en style Spirituel: \\
& German & ändert die Eingabe eines bestimmten Benutzers in einen Witziger Stil: \\
& Italian & Cambia la frase o il testo immesso da un utente in stile Spiritoso: \\
& Chinese & 将给定用户的输入句子或文本更改为机智风格: \\
& Korean & 주어진 사용자의 입력을 재치있는 문체로 변경한다: \\
& Japanese & 指定されたユーザーの入力文またはテキストを ウィットに富んだ スタイルに変更する: \\
\midrule
\multirow{8}{*}{Tone Adj. (Paraphrase)} & English & Paraphrase the following text: \\
& Spanish & Parafrasea el siguiente texto: \\
& French & Paraphraser le texte suivant: \\
& German & Fassen Sie den folgenden Text zusammen: \\
& Italian & Parafrasare il testo seguente: \\
& Chinese & 解释以下文字: \\
& Korean & 다음 텍스트를 의역하세요: \\
& Japanese & 次のテキストを言い換えてください: \\
\midrule
\multirow{8}{*}{Question Answering} & English & Answer the following question: \\
& Spanish & Responde a la siguiente pregunta: \\
& French & Réponds à la question suivante: \\
& German & Beantworten Sie die folgende Frage: \\
& Italian & Rispondi alla seguente domanda: \\
& Chinese & 回答以下问题: \\
& Korean & 다음 질문에 답하시오: \\
& Japanese & 次の質問に答えましょう: \\
\bottomrule
\end{tabular}
}
\end{center}
\caption{Prompts for each problem type and language, from \cite{ceritli2025hydraopt}.}
\label{tab:prompts_full}
\end{table*}
\end{CJK}

\end{document}